\documentclass{article}

\usepackage[utf8]{inputenc} 
\usepackage[T1]{fontenc}    
\usepackage{hyperref}       
\usepackage{url}            
\usepackage{booktabs}       
\usepackage{amsfonts}       
\usepackage{nicefrac}       
\usepackage{microtype}      
\usepackage{lipsum}
\usepackage{xcolor}
\usepackage[colorinlistoftodos]{todonotes}

\usepackage{fullpage}
\usepackage{amssymb}
\usepackage{amsmath}
\usepackage{amsthm}
\usepackage{authblk}
\usepackage{latexsym}
\usepackage{graphicx}
\usepackage{wrapfig}
\usepackage{color}
\usepackage{url}
\usepackage{algorithm}
\usepackage{algorithmic}
\usepackage{bbm}
\usepackage{blkarray}
\usepackage{comment}
\usepackage{bm}
\usepackage{mathtools}

\newcommand{\sq}{\textup{\textrm{sq}}}
\theoremstyle{definition}

\newtheorem{lemma}{Lemma}
\newtheorem{remark}{Remark}
\newtheorem{theorem}[lemma]{Theorem}

\usepackage[super]{nth}

\def\pd#1#2{{\frac{\partial #1}{\partial #2}}}

\newcommand*\samethanks[1][\value{footnote}]{\footnotemark[#1]}

\def\keywordname{{\emph Keywords:}}%
\def\keywords#1{\par\addvspace\medskipamount{\rightskip=0pt plus1cm
\def\and{\ifhmode\unskip\nobreak\fi\ $\cdot$
}\noindent\keywordname\enspace\ignorespaces#1\par}}

\def\mindex#1{\index{#1}}

\def\sq{\hbox{\rlap{$\sqcap$}$\sqcup$}}
\def\qed{\ifmmode\sq\else{\unskip\nobreak\hfil
\penalty50\hskip1em\null\nobreak\hfil\sq
\parfillskip=0pt\finalhyphendemerits=0\endgraf}\fi\medskip}

\long\def\defbox#1{\framebox[.9\hsize][c]{\parbox{.85\hsize}{%
\parindent=0pt
\baselineskip=12pt plus .1pt      
\parskip=6pt plus 1.5pt minus 1pt 
 #1}}}

\long\def\beginbox#1\endbox{\subsection*{}%
\hbox{\hspace{.05\hsize}\defbox{\medskip#1\bigskip}}%
\subsection*{}}

\def\endbox{}

\newsavebox{\junk}
\savebox{\junk}[1.6mm]{\hbox{$|\!|\!|$}}

\def\det{{\mathop{\rm det}}}

\def\bfmath#1{{\mathchoice{\mbox{\boldmath$#1$}}%
{\mbox{\boldmath$#1$}}%
{\mbox{\boldmath$\scriptstyle#1$}}%
{\mbox{\boldmath$\scriptscriptstyle#1$}}}}

\def\bfmY{\bfmath{Y}}

\def\bfmhhaY{\bfmath{\hhaY}} 
\def\bfmhhaY{\hbox to 0pt{$\widehat{\bfmY}$\hss}\widehat{\phantom{\raise 1.25pt\hbox{$\bfmY$}}}}

\def\til={{\widetilde =}}

\def\clF{{\cal F}}

\def\clL{{\cal L}}

\def\clN{{\cal N}}

 \def\FRAC#1#2#3{\genfrac{}{}{}{#1}{#2}{#3}}

\def\ddtp{{\mathchoice{\FRAC{1}{d^{\hbox to 2pt{\rm\tiny +\hss}}}{dt}}%
{\FRAC{1}{d^{\hbox to 2pt{\rm\tiny +\hss}}}{dt}}%
{\FRAC{3}{d^{\hbox to 2pt{\rm\tiny +\hss}}}{dt}}%
{\FRAC{3}{d^{\hbox to 2pt{\rm\tiny +\hss}}}{dt}}}}

\def\average#1,#2,{{1\over #2} \sum_{#1}^{#2}}

\def\eye(#1){{\bf(#1)}\quad}

\def\eq#1/{(\ref{e:#1})}

\newcommand{\beqn}[1]{\notes{#1}%
\begin{eqnarray} \elabel{#1}}

\newcommand{\eeqn}{\end{eqnarray} }

\newcommand{\beq}[1]{\notes{#1}%
\begin{equation}\elabel{#1}}

\newcommand{\eeq}{\end{equation}}

\def\bdes{\begin{description}}
\def\edes{\end{description}}

\newcounter{rmnum}

\newcounter{anum}

{\end{list}}

\def\ass(#1:#2){(#1\ref{#1:#2})}

\def\ritem#1{
\item[{\sf \ass(\current_model:#1)}]
}

\newenvironment{recall-ass}[1]{%
\begin{description}
\def\current_model{#1}}{
\end{description}
}

\long\def\comment#1{}

\newfont{\bbb}{msbm10 scaled 700}

\newfont{\bb}{msbm10 scaled 1100}

\renewcommand{\det}{{\hbox{det}}}

\title{Exact Gradient Computation for Spiking Neural Networks Through Forward Propagation}

\author[1]{Jane H. Lee\thanks{Equal contribution}}
\author[2]{Saeid Haghighatshoar\samethanks}
\author[3]{Amin Karbasi}

\affil[1]{Department of Computer Science, Yale University}
\affil[2]{SynSense, Zurich, Switzerland}
\affil[3]{Department of Electrical Engineering, Yale University}

\begin{document}
\maketitle

\begin{abstract}
  Spiking neural networks (SNN) have recently emerged as alternatives to traditional neural networks, owing to energy efficiency benefits and capacity to better capture biological neuronal mechanisms. However, the classic backpropagation algorithm for training traditional networks has been notoriously difficult to apply to SNN due to the hard-thresholding and discontinuities at spike times. Therefore, a large majority of prior work believes exact gradients for SNN w.r.t. their weights do not exist and has focused on approximation methods to produce surrogate gradients. In this paper, (1) by applying the implicit function theorem to SNN at the discrete spike times, we prove that, albeit being non-differentiable in time, SNNs have well-defined gradients w.r.t. their weights, and (2) we propose a novel training algorithm, called \emph{forward propagation} (FP), that computes exact gradients for SNN. FP exploits the causality structure between the spikes and allows us to parallelize computation forward in time. It can be used with other algorithms that simulate the forward pass, and it also provides insights on why other related algorithms such as Hebbian learning and also recently-proposed surrogate gradient methods may perform well.

\keywords{spiking neural networks \and exact gradients \and neuromorphic computation}

\end{abstract}

\section{Introduction}
While artificial neural networks have achieved state-of-the-art performance on various tasks, such as in natural language processing or computer vision, these networks are usually large, complex, and their computation consumes a lot of energy. Spiking neural networks (SNNs), inspired by biological neuronal mechanisms and sometimes referred to as the third generation of neural networks \cite{maass_third_gen}, have garnered considerable attention recently \cite{roy_snn_computing_2019, panda_snn_residual_2020, cao_snn_object_recognition_2015, comsa_temporal_coding_2020, diehl_unsupervised_snn_2015} as low-power alternatives. For instance, SNNs have been shown to yield 1-2 orders of magnitude energy saving over ANNs on emerging neuromorphic hardware \cite{true_north_2015, loihi_2018}. SNNs have other unique properties, owing to their ability to model biological mechanisms such as dendritic computations with temporally evolving potentials \cite{dendritic} or short-term plasticity, which allow them to even outperform ANNs in accuracy in some tasks \cite{plasticity}. The power of neuromorphic computing can even be seen in ANNs, e.g., \cite{jeffares2022spikeinspired} use rank-coding in ANN inspired by the temporal encoding of information in SNNs.
However, due to the discontinuous resetting of the membrane potential in spiking neurons, e.g., in Integrate-and-Fire (IF) or Leaky-Integrate-and-Fire (LIF) type neurons \cite{burkitt_integrate_fire_2006, kornijcuk_lif_2016}, 
it is notoriously difficult to calculate gradients and train SNNs by conventional methods. For instance, \cite{jeffares2022spikeinspired} use the fact that “spike coding poses difficulties...and training that require ad hoc mitigation” and “SNNs are particularly difficult to analyse mathematically” to motivate rank-coding for ANN. 
As such, many existing works on training SNN do so without exact gradients, which range from heuristic rules like Hebbian learning \cite{kempter_hebbian_1999, ruf_hebbbian_2006} and STDP \cite{lee_stdp_pretrain_2018, lobov_stdp_2020}, SNN-ANN conversion \cite{rueckauer_conversion_2017, Ding2021OptimalAC, ho_conversion_2021}, and surrogate gradient approximations \cite{Neftci2019SurrogateGL}.

In this work, by applying the implicit function theorem (IFT) at the firing times of the neurons in SNN, we first show that under fairly general conditions, gradients of loss w.r.t. network weights are well-defined. We do this by proving that the conditions for IFT are always satisfied at firing times. We then provide what we call a \emph{forward-propagation} (FP) algorithm which uses the causality structure in network firing times and our IFT-based gradient calculations in order to calculate exact gradients of the loss w.r.t. network weights. We call it forward propagation because intermediate calculations needed to calculate the final gradient are actually done forward in time (or forward in layers for feed-forward networks). We highlight the following features of our method:
\begin{itemize}
    \item Our method can be applied in networks with arbitrary recurrent connections (up to self loops) and is agnostic to how the forward pass is implemented. We provide an implementation for computing the firing times in the forward pass, but as long as we can obtain accurate firing times and causality information (for instance, using existing libraries), we can calculate gradients.
    
    \item Our method can be seen as an extension of Hebbian learning as it illustrates that the gradient w.r.t. a weight $W_{ji}$ connecting neuron $j$ to neuron $i$ is almost an average of the feeding kernel $y_{ji}$ between these neurons at the firing times. In the context of Hebbian learning (especially from a biological perspective), this is interpreted as the well-known fact that \textit{stronger feeding/activation amplifies the association between the neurons.} \cite{hebbian_overview_2013, gerstner_kistler_naud_paninski_2014}
    
    \item In our method, the smoothing kernels $y_{ji}$ arise naturally as a result of application of IFT at the firing times, resembling the smoothing kernels applied in surrogate gradient methods. As a result (1) our method sheds some light on why the surrogate gradient methods may work quite well and (2) in our method, the smoothing kernels $y_{ji}$ vary according to the firing times between two neurons; thus, they can be seen as an adaptive version of the fixed smoothing kernels used in surrogate gradient methods.
    
    \item Most of the methods in the literature apply a time-quantized version of the neuron dynamics and convert the continuous-time system into a discrete-time system. While we derive results in the continuous time regime, our IFT formulation is also applicable in these discrete-time scenarios. To do so, one needs to treat the weight parameters and all the time-quantized versions of the variables (such as synaptic and membrane potential, etc.) as separate variables. 
    The number of these state variables however grows proportionally to the simulation time and the precision of the time quantization, which is why the continuous-time regime is preferred.
\end{itemize}

\subsection{Related Work}
A review of learning in deep spiking networks can be found at \cite{tavanaei_overview_2018, pfeiffer_opportunities_2018, roy_snn_computing_2019, wang_review_2020}, with \cite{roy_snn_computing_2019} discussing also developments in neuromorphic computing in both software (algorithms) and hardware.  \cite{Neftci2019SurrogateGL} focuses on surrogate gradient methods, which  use smooth activation functions in place of the hard-thresholding for compatibility with usual backpropagation and have been used to train SNNs in a variety of settings \cite{esser_conv_2016, bellec_lstm_NEURIPS2018, huh_gd_NEURIPS2018, zenke_superspike_2018, shrestha_slayer_NEURIPS2018, safa_convsnn_2021}. 

A number of works explore backpropagation in SNNs \cite{spike_prop, macro_micro_backprop, spike_training_rsnn}. The SpikeProp \cite{spike_prop} framework assumes a linear relationship between the post-synaptic input and the resultant spiking time, which our framework does not rely on. The method in \cite{macro_micro_backprop} and its RSNN version \cite{spike_training_rsnn} are limited to a rate-coded loss that depends on spike counts. The continuous ``spike time'' representation of spikes in our framework is related to temporal coding \cite{temporal_coding}, but the authors of \cite{temporal_coding} in the context of differentiation of losses largely ignore the discontinuities that occur at spikes times, stating ``the derivative...is discontinuous at such points [but] many feedforward ANNs use activation functions with a discontinuous first derivative''. In contrast with \cite{temporal_coding}, we prove that exact gradients can be calculated despite this discontinuity. 

As mentioned in \cite{eventbasedexactgrad2021}, applying methods from optimal control theory to compute exact gradients in hard-threshold spiking neural networks has been recognized \cite{Selvaratnam2000LearningMO, kuroe_adjoint_2010, kuroe-synthesizing_2010}. However, unlike in our setting these works consider a neuron with a two-sided threshold and provide specialized algorithms for specific loss functions. Most related to our work is the recent EventProp \cite{eventbasedexactgrad2021} which derives an algorithm for a continuous-time spiking neural network by applying the adjoint method (which can be seen as generalized backpropagation) together with proper partial derivative jumps. EventProp calculates the gradients by accumulating adjoint variables while computing adjoint state trajectories via simulating another continuous-time dynamical system with transition jumps in a backward pass, but our algorithm computes gradients with just firing time and causality information. In particular, the only time we need to simulate continuous-time dynamics is in the forward pass. 

\section{Spiking Neural Networks}
In this section, we first describe the precise models we use throughout the paper for the pre-synaptic and pos-synaptic behaviors of spiking neurons. We then explain the dynamics of a SNN and the effects of spike generations. 
\subsection{Pre-Synaptic Model}
For the ease of presentation, a generic structure of a SNN is illustrated in Fig.\,\ref{fig:snn_model_pre_post} on the left. There are many different models to simulate the nonlinear dynamics of a spiking neuron (e.g., see \cite{gerstner_kistler_naud_paninski_2014}). In this paper, we adopt the Leaky-Integrate-and-Fire (LIF) model which consists of three main steps.

\subsubsection{Synaptic Dynamics} A generic neuron $i$ is stimulated through a collection of input neurons, its neighborhood $\clN_i$. Each neuron $j \in \clN_i$ has a synaptic connection to $i$ whose dynamics is modelled by a \nth{1}-order low-pass $RC$ circuit that smooths out the Dirac Delta currents it receives from neuron $j$. Since this system is linear and time-invariant (LTI), it can be described by its impulse response $$h^s_{j}(t)=e^{-\alpha_{j} t} u(t),$$ where  $\alpha_j=\frac{1}{\tau^s_{j}}$ and $
\tau^s_{j}=R^s_{j}C^s_{j}$ denotes the synaptic time constant of neuron $j$, and $u(t)$ denotes the Heaviside step function. Therefore, the output synaptic current $I_j(t)$ can be written as 
\begin{align}\label{eq:currentj}
    I_j(t)=h^s_{j}(t) \star \sum_{f\in \clF_j} {\delta(t-f)} = \sum_{f\in \clF_j} h^s_{j}(t-f),
\end{align}
where $\mathcal{F}_j$ is the set of output firing times from neuron $j$. Note that in Eq. \eqref{eq:currentj} we used the fact that convolution with a Direct Delta function $h^s_{j}(t)\star \delta(t-f)=h^s_{j}(t-f)$, is equivalent to shifts in time.

\subsubsection{Neuron Dynamics} The synaptic current of all stimulating neurons is weighted by $W_{ji}$, $j \in \clN_i$, and builds the weighted current that feeds the neuron. The dynamic of the neuron can be described by yet another \nth{1}-order low-pass $RC$ circuit with a time constant $\tau^n_{i}=R^n_{i}C^n_{i}$ and with an impulse response $h^n_{i}(t)=e^{-\beta_{i} t} u(t)$ where $\beta_{i}=\frac{1}{\tau^n_{i}}$. The output of this system is the membrane potential $V_i(t)$.

\subsubsection{Hard-thresholding and spike generation} The membrane potential $V_i(t)$ is compared with the firing threshold $\theta_i$ of neuron $i$ and a spike (a delta current) is produced by neuron when $V_i(t)$ goes above $\theta_i$. Also, after spike generation, the membrane potential is reset/dropped immediately by $\theta_i$ (reset to zero).

\subsection{Post-Synaptic Kernel Model}\label{sec:post-syn}
\begin{figure*}[t]
    \centering
    \includegraphics[width=\textwidth]{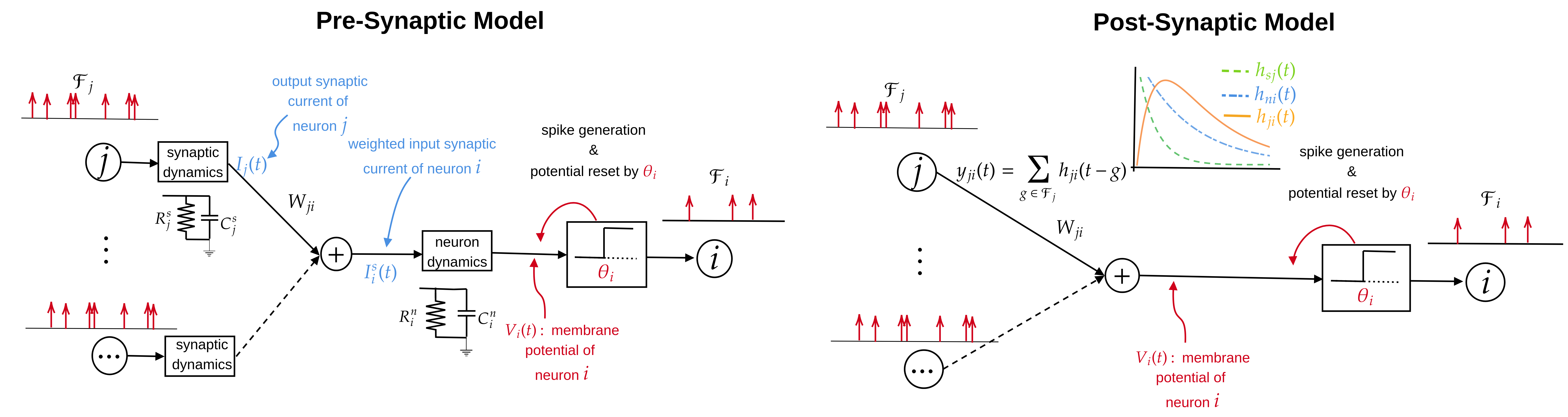}

    \caption{\textbf{(Left)} A generic structure of a spiking neural network: (i) spikes (train of Dirac Delta currents) $\clF_j$ coming from a generic input neuron $j$ pass through the synaptic RC circuit with a time constant $\tau^s_{j}=R^s_{j}C^s_{j}$ and build the synaptic current $I_j(t)$, (ii) synaptic current $I_j(t)$ are weighted by $W_{ji}$ and build the input current $\sum_{j} W_{ji} I_j(t)$, (iii) this current is filtered through neuron $i$ as an RC circuit with a time constant $\tau^n_{i}=R^n_{i}C^n_{i}$ and produces the membrane potential $V_i(t)$, (iv) membrane potential $V_i(t)$ is compared with the threshold $\theta_i$ and a current spike is produced when it passes above $\theta_i$, then (v) membrane potential is reset/dropped by $\theta_i$ immediately after the spike generation. \textbf{(Right)} Post-synaptic kernel model of the SNNs. In this model neuron $j \in \clN_i$ stimulates neuron $i$ through the smooth kernel $y_{ji}(t)=\sum_{g\in \clF_i} h_{ji}(t-g)$ rather than the abrupt spiking signal $\sum_{g\in \clF_j} \delta(t-g)$ as adopted in pre-synaptic model.}
    \label{fig:snn_model_pre_post}
\end{figure*}

We call the model illustrated in the left of Fig.\,\ref{fig:snn_model_pre_post} the pre-synaptic model, as the spiking dynamics of the stimulating neurons $\clN_i$ of a generic neuron $i$ appear before the synapse.
In this paper, we will work with a modified but equivalent  model in which we combine the synaptic and neuron dynamics, and consider the effect of spiking dynamics of $\clN_i$ directly on the membrane potential after it is being smoothed out by the synapse and neuron low-pass filters. 
We call this model the post-synaptic or kernel model of the SNN.

To derive this model, we simply use the fact that the only source of non-linearity in SNN is  hard-thresholding during the spike generation. And, in particular, SNN dynamics from the stimulating neuron $j \in \clN_i$ until the membrane potential $V_i(t)$ is completely linear and can be described by the joint impulse response 
\begin{align}
h_{ji}(t)&=h^s_{j}(t)\star h^n_{i}(t) \nonumber\\
&=\int_{-\infty}^\infty h^s_{j}(\tau)h^n_{i}(t-\tau) d\tau \nonumber\\
&= \int_0^ t e^{-\alpha_{j}\tau}  e^{-\beta_{i}(t-\tau)}  d\tau \nonumber\\
&= \frac{e^{-\alpha_{j}t} - e^{-\beta_{i}t} }{\beta_{i} - \alpha_{j}} u(t). \label{eq:combined_response}
\end{align}
Therefore the whole effect of spikes $\clF_j$ of neuron $j \in \clN_i$ on the membrane potential can be written in terms of kernel
\begin{align*}
    y_{ji}(t)= \sum_{f \in \clF_j} h_{ji}(t-f).
\end{align*}
We call this model post-synaptic since the effect of dynamic of neuron $j\in \clN_i$ on $V_i(t)$ is considered after being processed by the synapse and even the neuron $i$. 
Using the linearity and applying super-position for linear systems, we can see that the effect of all spikes coming for all stimulating neurons $\clN_i$, can be written as 
\begin{align}
    V_i^\circ(t)=\sum_{j\in \clN_i} W_{ji} y_{ji}(t),\label{eq:snn_lin_dyn}
\end{align}
where $W_{ji}$ is the weight from neuron $j$ to $i$. We used $V_i^\circ(t)$ to denote the contribution to the membrane potential $V_i(t)$ after neglecting the potential reset due to hard-thresholding and spike generation. 
Fig.\,\ref{fig:snn_model_pre_post} (right) illustrates the post-synaptic model for the SNN.

\begin{remark}
Our main motivation for using this equivalent model comes from the fact that even though the spikes are not differentiable functions, the effect of each stimulating neuron $j \in \clN_i$ on neuron $i$ is written as a well-defined and (almost everywhere) differentiable kernel $y_{ji}(.)$.
\hfill $\lozenge$
\end{remark}

\begin{remark}[Connection with the surrogate gradients]\label{rem:surrogate_grad}
Intuitively speaking, and as we will show rigorously in the following sections, the kernel model derived here immediately shows that SNNs have an intrinsic smoothing mechanism for their abrupt spiking inputs, through the low-pass impulse response $h_{ji}(t)$ between their neurons. As a result, one does not need to introduce any additional artificial smoothing to derive surrogate gradients by modifying the neuron model in the backward gradient computation path. We will use this inherent smoothing to prove that SNNs indeed have well-defined gradients. Interestingly, our derivation of the exact gradient based on this inherent smoothing property intuitively explains that even though surrogate gradients are not exact, they may be close to and yield a similar training performance as the exact gradients. \hfill $\lozenge$
\end{remark}

\subsection{SNN Full Dynamics}
In the post-synaptic kernel model, we already specified the effect of spikes from stimulating neurons as in \eqref{eq:snn_lin_dyn}. To have a full picture of the SNN dynamics, we need to specify also the effect of spike generation. The following theorem completes this.

\begin{theorem} \label{thm:response}
Let $i$ be a generic neuron in SNN and let $\clN_i$ be the set of its stimulating neurons.
Let $h^n_{i}(t)$ and $h^s_{j}(t)$ be the impulse response of the neuron $i$ and synapse $j\in \clN_i$, respectively, and let $h_{ji}(t)=h^n_{i}(t)\star h^s_{j}(t)$.  Then the membrane potential of the  neuron $i$ for all times $t$ is given by 
\begin{align}
    V_i(t) = V_i^{\circ}(t) - \sum_{f \in \mathcal{F}_i}\theta_i h^n_{i}(t -f), \label{eq:potential}
\end{align}
where $y_{ji}(t)=\sum_{g \in \clF_j} h_{ji}(t-g)$ denotes the smoothed kernel between the neuron $i$ and $j \in \clN_i$, and $\theta_i$ denotes the spike generation threshold of the neuron $i$. \hfill $\qedsymbol$
\end{theorem}

\begin{proof}
In the following, we provide a a simple and intuitive proof. An alternative and more rigorous proof by induction on the number of firing times of neuron $i$ is provided in the Appendix \ref{appendix:alt_proof_thm3}.

\noindent {\bf Proof (i)}: We use the following simple result/computation-trick from circuit theory that in an RC circuit, abrupt dropping of the potential of the capacitor by $\theta_i$ at a specific firing time $f\in \clF_i$ can be mimicked by adding a voltage source $-\theta_i u(t-f)$ series with the capacitor. If we do this for all the firing times of the neuron, we obtain a linear RC circuit with two inputs: (i)\,weighted synaptic current coming from the neurons $\clN_i$, (ii)\,voltage sources $\{-\theta_i  u(t-f): f \in \mathcal{F}_i\}$. This is illustrated in Fig.\,\ref{fig:delta_circuit_model}.

\begin{figure}[tb]
    \centering
    \includegraphics[width=0.73\textwidth]{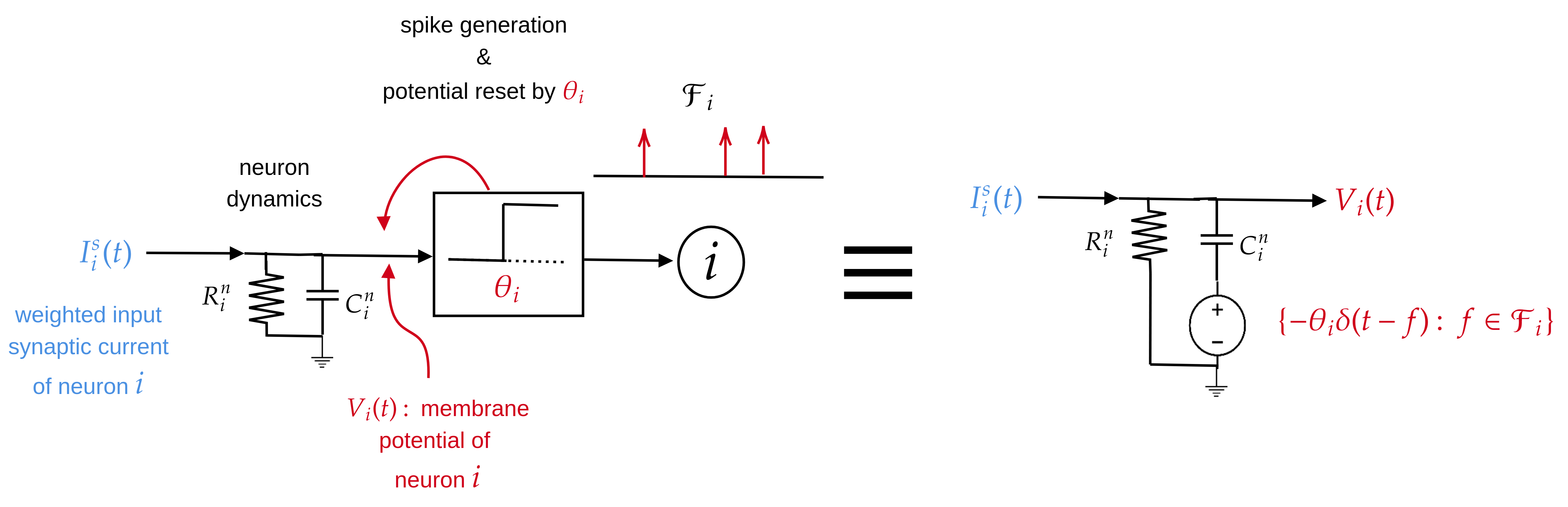}
    \includegraphics[width=0.25\textwidth]{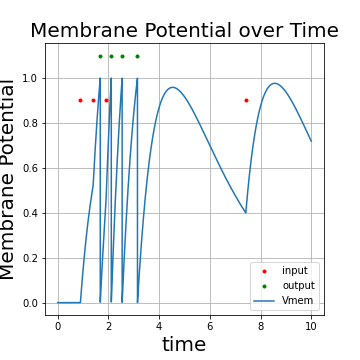}
    \caption{\textbf{(Left)} Equivalence of response for: (i)\,a nonlinear neuron with weighted synaptic currents $I(t)$ and spike generation, and (ii)\,a linear neuron with input $I(t)$ and Heaviside voltages $\{-\theta_i u(t-f): f\in \clF_i\}$. \textbf{(Right)} Example membrane potential over time using Eq. \eqref{eq:potential_expanded}.}
    \label{fig:delta_circuit_model}\label{fig:mem_potential}
\end{figure}

The key observation is that although this new circuit is obtained after running the dynamics of the neuron and observing its firing times $\clF_i$, as far as the membrane potential $V_i(t)$ is concerned, the two circuits are equivalent. 
Interestingly, after this modification, the new circuit is a completely linear circuit and we can  apply the super-position principle for linear circuits to write the response of the neuron as the summation of: (i)\,the response $V_i^{(1)}(t)$ due to the weighted synaptic current $I^s_i(t)$ in the input (as in the previous circuit), and (ii)\,the response $V_i^{(2)}(t)$ due to Heaviside voltage sources $\{-\theta_i u(t-f): f \in \mathcal{F}_i\}$. From \eqref{eq:snn_lin_dyn}, $V_i^{(1)}(t)$ is simply given by 
\begin{align*}
    V_i^{(1)}(t)=\sum_{j \in \clN_i} W_{ji} y_{ji}(t).
\end{align*}
The response of an RC circuit to a Heaviside voltage function $-\theta_i u(t-f)$ is given by $-\theta_i h^n_{i}(t-f)$ where $h^n_{i}(t)$ is the impulse response of the neuron $i$ as before. We also used the time invariance property (for shift by $f$) and a well-known result from circuit theory (Thevenin-Norton theorem) that for an RC circuit the impulse response due to a Delta current source is the same as the impulse response due to a Heaviside voltage source. 
The response to all Heaviside voltage functions, from super-position principle, is simply given by
\begin{align*}
    V_i^{(2)} (t)= -\theta_i\sum_{f \in \mathcal{F}_i} h^n_{i}(t-f).
\end{align*}
Therefore, we obtain that
\begin{align}
V_i(t) &= V_i^{(1)} (t) + V_i^{(2)} (t) \nonumber\\
&= \sum_{j\in \mathcal{N}_i} W_{ji} y_{ji}(t) - \sum_{f \in \mathcal{F}_i}\theta_i h^n_{i}(t -f). \label{eq:potential_expanded}
\end{align}
This completes the proof. See Fig.\,\ref{fig:mem_potential} for an illustration.
\end{proof}

\section{Exact Gradient Computation via Implicit Function Theorem}\label{sec:exact_grad}
In this section, we introduce the Implicit Function Theorem (IFT) which will be our main tool for proving the existence of gradients for SNNs. We state the theorem here for the reader's convenience (and some examples of why one needs IFT for certain problems are provided in the Appendix \ref{appendix:ift}).

\begin{theorem}[Implicit Function Theorem]
Let $\phi: \mathbb{R}^n \times \mathbb{R}^m \to \mathbb{R}^m$ be a differentiable function and let $\mathcal{Z}=\{(x,y) \in \mathbb{R}^n \times \mathbb{R}^m: \phi(x,y)=0\}$ be the zero-set of $\phi$. Suppose that $\mathcal{Z} \not = \emptyset$ and let $(x_0,y_0) \in \mathcal{Z}$ be an arbitrary point. Also, let $\frac{\partial \phi}{\partial y} \phi(x_0,y_0)$ be the $m\times m$ matrix of partial derivatives w.r.t. $y$ and assume that it is non-singular, i.e., $\det\big ( \frac{\partial \phi}{\partial y} (x_0,y_0) \big ) \not = 0$. Then, 
\begin{itemize}
    \item There is an open neighborhood $\mathcal{N}_x$ around $x_0$ and an open neighborhood $\mathcal{N}_\mathrm{y}$ around $y_0$ such that $\frac{\partial \phi}{\partial y} \phi (x,y)$ is non-singular for all $(x,y) \in \mathcal{N}:=\mathcal{N}_\mathrm{x} \times \mathcal{N}_\mathrm{y}$ (including of course the original $(x_0,y_0)$.
    
    \item There is a function $\psi: \mathcal{N}_\mathrm{x} \to \mathcal{N}_\mathrm{y}$ such that $(x,\psi(x))$ belongs to the zero set $\mathcal{Z}$, namely, $\phi(x,\psi(x))=0$, for all $x \in \mathcal{N}_\mathrm{x}$; therefore, the variables $y$ in $\mathcal{N}_\mathrm{y}$ can be written as a function $y=\psi(x)$ of the variables $x$ in $\mathcal{N}_\mathrm{x}$.
    
    \item $\psi$ is a differentiable function of $x$ for $x \in \mathcal{N}_\mathrm{x}$ and 
    \begin{align*}
        \frac{\partial \phi}{\partial y}  \times \frac{\partial \psi}{\partial x} + \frac{\partial \phi}{\partial x}=0,
    \end{align*}
    which from the non-singularity of $\frac{\partial \phi}{\partial y}$ yields
    \begin{align*}
         \frac{\partial \psi}{\partial x} = - \Big ( \frac{\partial \phi}{\partial y} \Big ) ^ {-1} \times \frac{\partial \phi}{\partial x}.
    \end{align*}
    
\end{itemize}

\end{theorem}

\subsection{Loss Formulation in SNNs}
To apply the IFT to SNNs, we need to specify the loss function we use for training such networks. Here, we consider a quite generic loss function of the form 
\begin{align}
    \clL= \ell_{\mathcal{F}}(\mathcal{F}; W) + \int_{0}^T \ell_V(V_o(t), \mathcal{F}; W) \textnormal{dt}, \label{eq:ell}
\end{align}
where  $\ell_\clF$ and $\ell_V$ are assumed to be differentiable functions of all their arguments, with $\ell_{\mathcal{F}}$ the part of the loss that depends on firing times $\clF=\sqcup_i \clF_i$ (disjoint union), and $\ell_{V}$ the part that depends on membrane potential at the output layer, respectively. 
Note that the second term $\ell_V(V_o(t), \mathcal{F}; W)$ is typically relevant in regression tasks where in those cases, we always assume that the output layer is linear without any firing and potential reset. The first term, in contrast, typically happens in classification tasks.

\begin{theorem}\label{thm:loss}
Let $\clL$ be the generic loss function as defined before in \eqref{eq:ell}. Then,
\begin{enumerate}
    \item[(i)] loss $\clL$ depends only on the spike firing times $\clF$ and the weights $W$, i.e., $\clL=\clL(\clF, W)$,
    \item [(ii)] $\clL(\clF,W)$ is a differentiable function of $\clF$ and $W$ if $\ell_V(V_o(t), \clF;W)$ and $\ell_{\clF}(\clF;W)$ are differentiable functions of all their arguments $(V_o(t), \clF; W)$,
    \item [(iii)] loss $\clL$ has well-defined gradients w.r.t. the weights $W$ if the spike firing times $\clF$ are differentiable w.r.t. the weights $W$. 
\end{enumerate}
\end{theorem}
\begin{proof}
(i) Note that in our post-synaptic kernel model derived in Section \ref{sec:post-syn}, the membrane potential of the output layer $V_o(t)$ can be written (in a more expanded form) as
\begin{align}
    V_o(t)=\sum_{j\in \clN_o}W_{jo} \sum_{g \in \clF_j} h_{jo}(t-g).\label{eq:linear}
\end{align}
Note that we dropped the term $- \theta_o \sum_{f\in \clF_o} h^n_{o}(t-f)$ due to potential reset because we always assume that the output neuron is linear in regression tasks where $V_o(t)$ appears directly in the loss. It is also seen that $V_o(t)$ at each time $t$ is a function of all the firing times $\clF$ and also weights $W$. 

(ii)\,Since $\ell_\clF$ is assumed to be a differentiable function of $\clF$ and $W$, we need to verify only the differentiability of the integral expression in \eqref{eq:ell}. 
First note that $h_{jo}(t)$ is a differentiable function except at $t=0$ where, albeit being non-differentiable, it has finite left and right derivatives. This implies that $V_o(t)$ in \eqref{eq:linear} is  differentiable at all $t$ except at the firing times of its stimulating neuron $\clN_o$, where at those points it has finite left and right derivatives. 
Therefore, we may write
\begin{align*}
    \frac{\partial}{\partial \clF} \int_{0}^T \ell_V(V_o(t), \mathcal{F}; W) \textnormal{dt}
    &= \int_{0}^T \pd{\ell_V}{V_o} (V_o(t), \mathcal{F}; W)  \pd{V_o(t)}{\clF} \\
    &+ \int_{0}^T \pd{\ell_V}{\clF}(V_o(t), \mathcal{F}; W) \textnormal{dt}. 
\end{align*}
Since $\ell_V$ is assumed to be a differentiable function of $\clF$, the second integral is well-defined. Also, $\ell_V$ is differentiable with respect to $V_o$. And $V_o(t)$, being a weighted combination of terms $h_{ji}(t-g)$ with $g \in \sqcup_{j \in \clN_o} \clF_j$, is a differentiable function of firing times $\clF$  except perhaps at finitely many points $t \in  \sqcup_{j \in \clN_o} \clF_j$ where at those points it may be discontinuous but has finite left and right derivatives. This implies that the first integral is also well-defined.

(iii) Since from (ii), the loss $\clL=\clL(\clF;W)$ is a differentiable function of both $\clF$ and $W$, we have that
\begin{align}
\pd{\clL}{W}=\clL_1 \pd{\clF}{W} + \clL_2 \label{eq:loss_chain_r}
\end{align}
where $\clL_1$ and $\clL_2$ denote the partial derivative of $\clL$ w.r.t. its \nth{1} and \nth{2} argument, and where we used the fact that from (ii) both $\clL_1$ and $\clL_2$ are well-defined. It is seen that the gradients of loss w.r.t. $W$ exist provided that the firing times $\clF$ are differentiable w.r.t. the weights.
This completes the proof.
\end{proof}
Theorem \ref{thm:loss} implies that to prove the existence of the gradients w.r.t. to the weights, which is needed for training the SNN, it is sufficient to prove that the firing times $\clF$ are differentiable w.r.t. the weights $W$. We will prove this in the next section by applying the IFT.

\subsection{Differentiability of Firing Times w.r.t. Weights}
Let us consider the set of equations for firing times by using \eqref{eq:potential_expanded}:
\begin{align}
    V_i(f) = \sum_{j \in \mathcal{N}_i} W_{ji} y_{ji} (f) - \theta_i \sum_{m<f} h^n_{i}(f - m) - \theta_i=0 \label{ift_eqs_rep}
\end{align}
where with some abuse of notation we use $f$ both for the firing time and its label $(i, f) \in \mathcal{F}=\sqcup_l \mathcal{F}_l$. We can write the equations for all the firing times as $\mathbb{V} (\mathcal{F}, W) = \mathbf{0}$ where $\mathbb{V}: \mathbb{R}^F \times \mathbb{R}^W \to \mathbb{R}^F$ is the nonlinear mapping connecting the $F$ firing times and $W$ weight parameters.

\begin{theorem}\label{perm_ift}
Let $\mathbf{P}$ be a permutation matrix sorting the firing times in $\mathcal{F}$ in an ascending order. Then, $\frac{\partial \mathbb{V}}{\partial \mathcal{F}} = \mathbf{P}^T \mathbf{L} \mathbf{P}$ where $\mathbf{L}$ is an $F \times F$ lower triangular matrix. Moreover, $\mathbf{L}$ has strictly positive diagonal elements $\mathbf{L}_{kk} >0$.
\end{theorem}
\begin{proof}
We note that due to causality (future firing times cannot affect past ones), the equation corresponding to a specific firing time $f\in \mathcal{F}$ can only have contribution from firing times less than $f$. In other words, $\frac{\partial V_f}{\partial g}=0$ for all $g < f$. Letting $\mathbf{P}$ be the permutation matrix sorting the firing times, therefore, the Jaccobian matrix of the sorted firing times given by $\mathbf{P} \frac{\partial \mathbb{V}}{\partial \mathcal{F}} \mathbf{P}^T$ should be a lower triangular matrix $\mathbf{L}$. This yields the first part $\frac{\partial \mathbb{V}}{\partial \mathcal{F}} = \mathbf{P}^T \mathbf{L} \mathbf{P}$. To check the second part, let $k$ be the index of a specific firing time $f$ in the sorted version. Let us denote the neuron corresponding to the firing $f$ by $i$. 
Then, we have that 
\begin{align*}
    \mathbf{L}_{kk}&=\frac{\partial \mathbb{V}_f}{\partial f}
    = \frac{d}{d f} V_i(f) \Big | _\text{\tiny all other firing times fixed}
    = V_i'(t) \Big | _{t=f^-} > 0
\end{align*}
which is equal to the left time derivative the potential $V_i(t)$ when it passes through the threshold $\theta_i$ at time $t=f$. It is worthwhile to mention that that since $V_i(f)$ is a differentiable function of $f$, it has both left and right derivatives and they are equal. However, this derivative is equal to only the left derivative of the potential. Note that this derivative should be strictly positive otherwise the potential will not surpass the firing threshold $\theta_i$ and no firing time will happen. This completes the proof.
\end{proof}
We will use the consequence of Theorem \ref{perm_ift} to always fulfill the conditions of the implicit function theorem (Theorem \ref{perm_ift2}), which will give us explicit formulas for the gradients of the network firing times w.r.t. network weights (Theorem \ref{thm:grad_exist}). 
\begin{theorem}\label{perm_ift2}
Let $\mathbb{V}(\mathcal{F}, W)=\mathbf{0}$ be the set of equations corresponding to the firing times. Then the $F\times F$ Jacobian matrix $\frac{\partial \mathbb{V}}{\partial \mathcal{F}}$ is non-singular. Moreover, the firing times $\mathcal{F}$ can be written as a differentiable function of the weights $W$.
\end{theorem}
\begin{proof}
The first part result follows from Theorem \ref{perm_ift}:
\begin{align*}
    \det\bigg(\frac{\partial \mathbb{V}}{\partial \mathcal{F}}\bigg) &= \det (\mathbf{P}^T \mathbf{L} \mathbf{P})= \det(\mathbf{P}) \det(\mathbf{L}) \det(\mathbf{P}^T)\\
    &= \det(\mathbf{L}) = \prod_{k} \mathbf{L}_{kk} >0,
\end{align*}
where we used the fact that $\det(\mathbf{P})=1$ for any permutation matrix $\mathbf{P}$. 
The second part follows from Implicit Function Theorem: $\mathbb{V}(\mathcal{F}, W)$ is a differentiable function of the firing times and weights and $\frac{\partial \mathbb{V}}{\partial \mathcal{F}}$ is non-singular, thus, firing times $\mathcal{F}$ can be written as a differentiable function of the weights.
\end{proof}

\begin{remark}
Using Theorem \ref{perm_ift} and \ref{perm_ift2} and applying the IFT, we have that
\begin{align*}
    \frac{\partial \mathbb{V}}{\partial \mathcal{F}} \times \frac{\partial \mathcal{F}}{\partial W} = - \frac{\partial \mathbb{V}}{\partial W}.
\end{align*}
After suitable sorting of the firing times $\mathcal{F}$ (thus, setting the required permutation matrix $\mathbf{P}$ to the identity matrix), this can be written as 
\begin{align}
    \mathbf{L} \frac{\partial \mathcal{F}}{\partial W} = - \frac{\partial \mathbb{V}}{\partial W}, \label{eq:ift_solve}
\end{align}
where $\mathbf{L}$ is a lower diagonal matrix. As a result, one can solve for the derivatives $\frac{\partial \mathcal{F}}{\partial W}$ recursively, so no matrix inversion is needed. \hfill $\lozenge$
\end{remark}
\begin{remark}
Our results hold for both feed-forward and recurrent networks since it is derived using only the causality relation between the firing times.
\hfill $\lozenge$
\end{remark}
\begin{remark}
The matrix $\frac{\partial \mathbb{V}}{\partial W}$ depends only on the values of kernels at the firing times. More specifically, let $f$ be a firing times of neuron $i$ and let $j \in \mathcal{N}_i$ be one of the feeding neurons of neuron $i$. Then, $\frac{\partial \mathbb{V}(f)}{\partial W_{ji}}= y_{ji}(f)$. 
Moreover, $\frac{\partial \mathbb{V}(f)}{\partial W_{kl}}=0$ if $l \not = i$ or $k \not \in \mathcal{N}_i$. \hfill $\lozenge$
\end{remark}

\begin{theorem}(Existence of gradients w.r.t. weights)\label{thm:grad_exist}
Let $\clL$ be a generic loss function for training a SNN as in \eqref{eq:ell} with $\ell_V$ and $\ell_\clF$ being differentiable w.r.t. their arguments. Then, $\clL$ has well-defined gradients w.r.t. weights.
\end{theorem}
\begin{proof}
    From Theorem \ref{thm:loss}, $\clL$ has well-defined gradients w.r.t. weights if the firing times $\clF$ as differentiable w.r.t. weights, which follows from Theorem \ref{perm_ift2} by applying the IFT. This completes the proof.
\end{proof}

\subsection{Generalization}
In this paper, we presented our results in the context of exponential kernels (also to be able to compare with \cite{eventbasedexactgrad2021}) where we showed that the response of the neuron membrane potential to the input and output spikes can be represented with the exponential feeding and refractory kernels $h_{ji}(t) = \frac{e^{-\alpha_{j}t} - e^{-\beta_{i}t} }{\beta_{i} - \alpha_{j}} u(t)$ and $-\theta_i h_i(t) = -\theta_i e^{-\alpha_{j}t} u(t)$. The more generic model for the neuron is the Spike Response Model (SRM) \cite{srm} where the membrane potential and output spikes can be written as
\begin{align*}
    V_i(t) = K^{in}_i(t) \star \sum_{j \in \clN_i} W_{ji} s^{in}_j(t) + K^{ref}_i(t) \star s^{out}_i(t), \\
    s^{out}_i(t)= u(V_i(t) - \theta_i)
\end{align*}
where $s^{in}_j(t)$ and $s^{out}_i(t)$ denote the input and output spikes and where $\theta_i$ is the firing threshold.
Our method based on IFT is still applicable as far as $K^{in}_i(t)$ and $K^{ref}_i(t)$ are differentiable functions.
Also, we need the additional condition that $K^{in}_i(0^+)=0$ to avoid sudden jumps due to the input spikes so that we can still write the membrane potential at any firing time $f$ as the equality condition
\begin{align}
\begin{split}
    V_i(f) &= \sum_{j \in \clN} W_{ji} \sum_{g \in \clF_j} K^{in}_i(f - g) - \sum_{e \in \clF_i: e < f} K^{ref}_i(f - e)= \theta_i.
\end{split}
\label{general_srm}
\end{align}
These conditions are definitely satisfied for $K^{in}_j(t)=h_{ji}(t)$ and $K^{out}_i(t)=-\theta_i h_i(t)$.
By applying the IFT to the differentiable equations \eqref{general_srm} corresponding to all the spike firing times, we can find the gradient of the firing times w.r.t. to the weight parameters.

\section{Implementation}\label{sec:implementation}
\subsection{Causality Graph}

Due to the formula in Eq. (\ref{eq:potential}), calculating the membrane potential at any given time just relies on keeping track of which firing times from the previous (feeding) neuron(s) caused the current one to spike. Thus to efficiently calculate partial derivatives, we will keep track of this information while calculating network firing outputs. A detailed explanation on a small example is given in \ref{appendix:causal_ex}. 

\subsection{Forward spike time computation}
Simulating an SNN in the forward pass and computing the firing times of its neurons requires solving the Euler integration corresponding to the differential equation of the synapse and membrane potentials. This is usually done approximately by quantizing time into small steps and iteratively updating potentials. There are several libraries such as snnTorch \cite{snntorch} that implement this. Our method for gradient computation can also use these methods where the firing times are computed.

Here, we propose another method that uses the impulse response (kernel) representation of the corresponding differential equations derived in (\ref{eq:combined_response}) and (\ref{eq:potential}) to compute the firing times exactly without any need for time quantization. The main idea behind this method is that for exponential synaptic and membrane impulse responses, one can always write the membrane potential of a neuron over a time interval $[t_0,t_1]$ at which the neuron receives no spikes at its input as $A e^{-\alpha t} + B e^{-\beta t}$ where $A$ and $B$ are some suitable coefficients and where $\alpha, \beta$ are the inverse synaptic and membrane time constants (common to all neurons), respectively.\footnote{For example, consider only two input spikes at times $t_1$ and $t_2 > t_1$ with associated weights $W_{1i}$ and $W_{2i}$. Then the total kernel value at $t \in [t_2, \infty)$ (at which there are no other input spikes) is given by $W_{1i} h_{1i}(t - t_1) + W_{2i} h_{2i}(t - t_2) = \frac{W_{1i} e^{\beta t_1} + W_{2i} e^{\beta t_2}}{\alpha-\beta} e^{-\beta t} + \frac{W_{1i} e^{\alpha t_1} + W_{2i} e^{\alpha t_2}}{\beta-\alpha} e^{-\alpha t} $. In case the neuron fires, e.g., at time $t_f$, we need to account for the potential resets by subtracting the term $\theta e^{-\beta(t - t_f)}u(t-t_f)$, which is again in the exponential form $\theta e^{\beta t_f} \times e^{-\beta t}$ for $t>t_f$. Thus the whole expression, for $t> t_2$ and before the next firing time, can be written as $Ae^{-\alpha t} + Be^{-\beta t}$.}
Thus the next firing time can be found by computing the time $t$, if there is any, at which this curve intersects the horizontal line $\theta$. Once this firing time is computed, we update $A$, $B$ and the search interval $[t_0, t_1]$ depending on whether the neuron receives any spikes before this firing time, and so on. This is summarzied in Algorithm \ref{alg:firing_time}.

\begin{remark}
Note that one can calculate partial derivatives immediately after solving for the firing time and computing the causality graph. In feed-forward networks, these calculations for neurons in the same layer can be done in parallel since the firing times of neurons in the same layer will not affect each other. 
\end{remark}

\begin{minipage}{0.47\textwidth}
\begin{algorithm}[H]
\caption{Firing Time Computation}
  \label{alg:firing_time}
\begin{algorithmic}
  \STATE {\bfseries Input:} Firing times $\clF = \sqcup_j \clF_j$ from neighbors $j \in \clN_i$ and weights $W_{ji}$. Hyperparameters $\alpha, \beta, \theta_i$.
   \STATE Initialize $A, B = 0, t_{ref} = 0$.
   \STATE Initialize empty queue.
   \FOR{$f$ (sorted) {\bfseries in} $\clF$ (where $f$ from neighbor $j$)}
   \STATE \begin{itemize}
       \item Append $f$ to queue. 
       \item Update $A \leftarrow A\cdot e^{-\alpha(f-t_{ref})} + W_{ji}/(\beta-\alpha)$ and $B \leftarrow B\cdot e^{-\beta(f-t_{ref})} + W_{ji}/(\alpha-\beta)$.
       \item Update $t_{ref} \leftarrow f$.
       \item Solve for $t$: $A e^{-\alpha t} + B e^{-\beta t} = \theta_i$. Add $t$ to output firing times.
       \item Update $A \leftarrow A\cdot e^{-\alpha(t - t_{ref})}$ and $B \leftarrow e^{-\beta(t - t_{ref})} - \theta_i$.
       \item Update $t_{ref} \leftarrow t$.
   \end{itemize}
   \STATE Add entire queue as causal edges to $t$.
   \ENDFOR
   \STATE {\bfseries Return} Causal graph and firing times.
\end{algorithmic}
\end{algorithm}
\end{minipage}
\hfill
\begin{minipage}{0.53\textwidth}
\begin{algorithm}[H]
  \caption{Forward Propagation}
   \label{alg:fp}
\begin{algorithmic}
   \STATE {\bfseries Input:} Network output firing times $\clF = \sqcup_i \clF_i$ for all $i$ and causal graph (e.g., by Alg. \ref{alg:firing_time}). Hyperparameters for network and loss.
   \STATE Initialize matrices $\mathbf {L}$, $\frac{\partial \mathcal{F}}{\partial W}$, and $\frac{\partial \mathbb{V}}{\partial W}$.
   \FOR{$f$ (sorted) {\bfseries in} $\clF$}
   \STATE Calculate partial derivatives of the firing time equation for $f$ output by neuron $i$: $V_i(f)-\theta_i=0$.
  \begin{itemize}
      \item Use causal information and Equation (\ref{eq:potential}) to fully describe $V_i(f)$.
      \item {\bfseries Update $\mathbf{L}$.}
      Calculate $\frac{\partial}{\partial f_{j \rightarrow i}} (V_i(f) - \theta_i)$ for each $f_{j \rightarrow i}$ in the causal graph for $f$. 
      \item {\bfseries Update $\mathbf{L}$.} Calculate $\frac{\partial}{\partial f} (V_i(f) - \theta_i)$. 
      \item {\bfseries Update $\frac{\partial \mathbb{V}}{\partial W}$.}
      Calculate $\frac{\partial}{\partial W_{j i}} (V_i(f) - \theta_i)$ for all weights $W_{j i}$ attached to neuron $i$. 
      \item {\bfseries IFT Step.}
      Solve Equation (\ref{eq:ift_solve}) via back substitution to update $\frac{\partial \mathcal{F}}{\partial W}$.
  \end{itemize}
   \ENDFOR
   \STATE Calculate $\frac{\partial \mathcal{L}}{\partial W}$ using final $\frac{\partial \mathcal{F}}{\partial W}$ via Eq. (\ref{eq:loss_chain_r}). 
\end{algorithmic}
\end{algorithm}
\end{minipage}

\subsection{Forward propagation for gradient computation}
The forward propagation algorithm (Algorithm \ref{alg:fp}) emerges from the earlier presented theorems and observations. We can derive partial derivatives of the total loss by calculating the partial derivatives of the network firing times w.r.t. network weights, which are in turn calculated by applying the implicit function theorem with appropriate partial derivatives of the equations that describe the membrane potentials at each firing time.

Again, due to the lower triangular structure of matrix $\mathbf{L}$ (see, e.g., Theorem \ref{perm_ift}), we can iteratively solve the linear system \eqref{eq:ift_solve} of IFT equations without having to do a full matrix inversion. 
This incurs a cost of $O(|\mathcal{F}|^2|\mathcal{W}|)$ in time, using $(1 + 2 + 3 + ... + |\mathcal{F}|)\times($ up to $|\mathcal{W}|)$ operations to solve for the $|\mathcal{F}| \times |\mathcal{W}|$ Jaccobian matrix. The memory cost is $O(|\mathcal{F}||\mathcal{W}|)$ to store the solutions and one of the Jacobians, where $O(|\mathcal{F}||\mathcal{W}|)$ is always needed for storing the gradients.

\section{Simulation}\label{sec:simulation}
Additional details on experiments presented in this section can be found in \ref{sec:experiment_details}. 

\subsection{XOR Task}

\begin{figure}
    \centering
    \includegraphics[width=0.9\columnwidth]{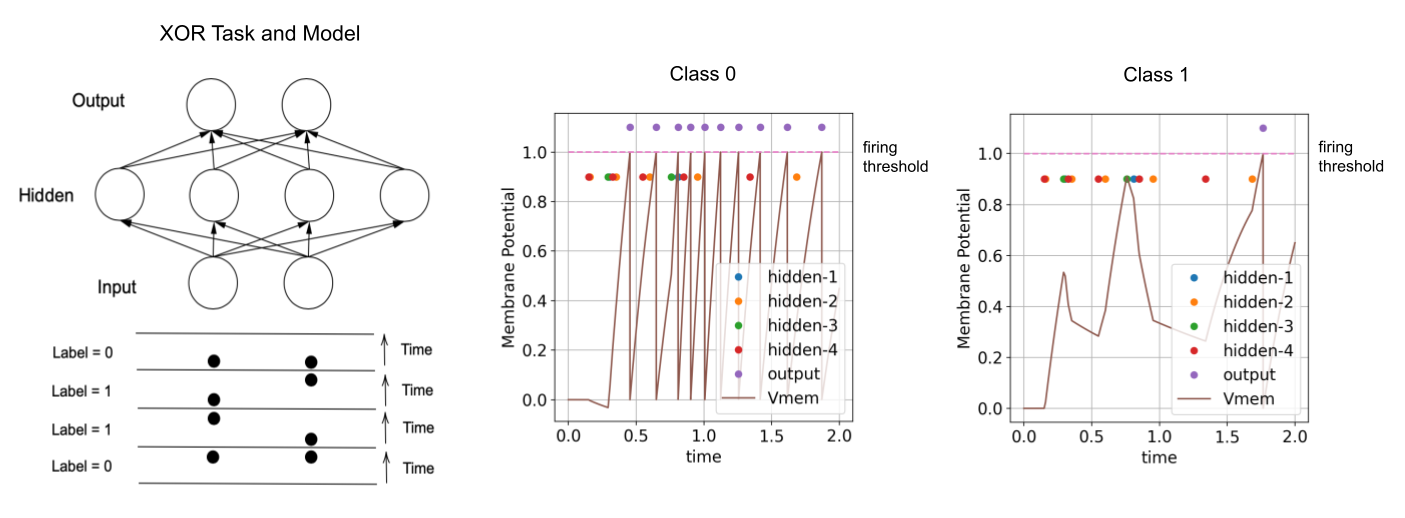}
    \caption{\textbf{(Left)} Model for XOR task. \textbf{(Right)} Given the input $(0, 0)$, output neurons have different voltage traces. Note that each output neuron has the same input firing times, from each of the 4 hidden layer neurons, but the network is able to learn weights that push the output neuron corresponding to label '1' to spike later, and the one corresponding to label '0' (true label) to spike earlier.}
    \label{fig:xor}
\end{figure}

To investigate whether the network can robustly learn to solve the XOR task as in \cite{temporal_coding}, we reproduced most of the experiment settings in \cite{temporal_coding} by coding each of the input spikes as 0.0 (early spike) or 2.0 (late spike), which feed into 4 hidden neurons, which in turn feed into 2 output neurons. We use a cross-entropy loss based on first spike times of the output neurons (so the label neuron should fire sooner than the other). For each of 1000 different random weight initializations, we trained until convergence with learning rate 0.1. Unlike in \cite{temporal_coding}, we consider one iteration of training to be just 1 full batch, rather than 100. Across all 1000 trials, the maximum steps to converge was 98, with the average being 17.52 steps. Compare this to maximum 61 training iterations (each iteration seeing 100 full batches of the four input patterns), with average 3.48 iterations in \cite{temporal_coding}. Figure \ref{fig:xor} illustrates the model implementing the XOR task, as well as a post-training simulation of the output neurons' membrane potentials for input $(0,0)$. 

\subsection{Iris Dataset}
\begin{figure}
    \centering
    \includegraphics[width=0.6\columnwidth]{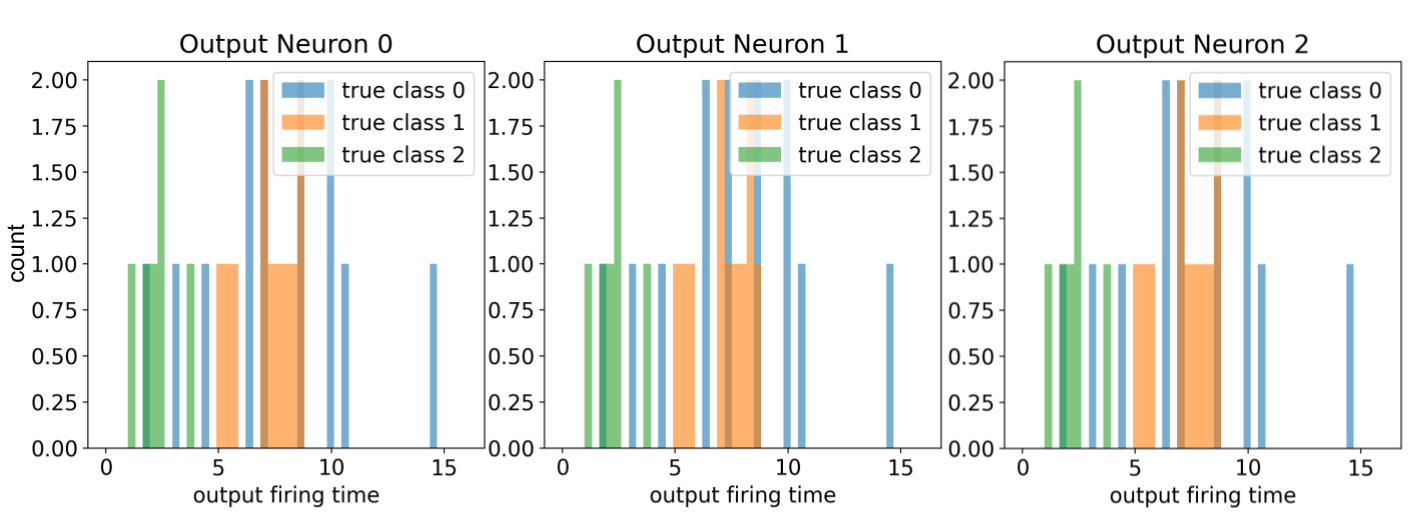}
    \includegraphics[width=0.6\columnwidth]{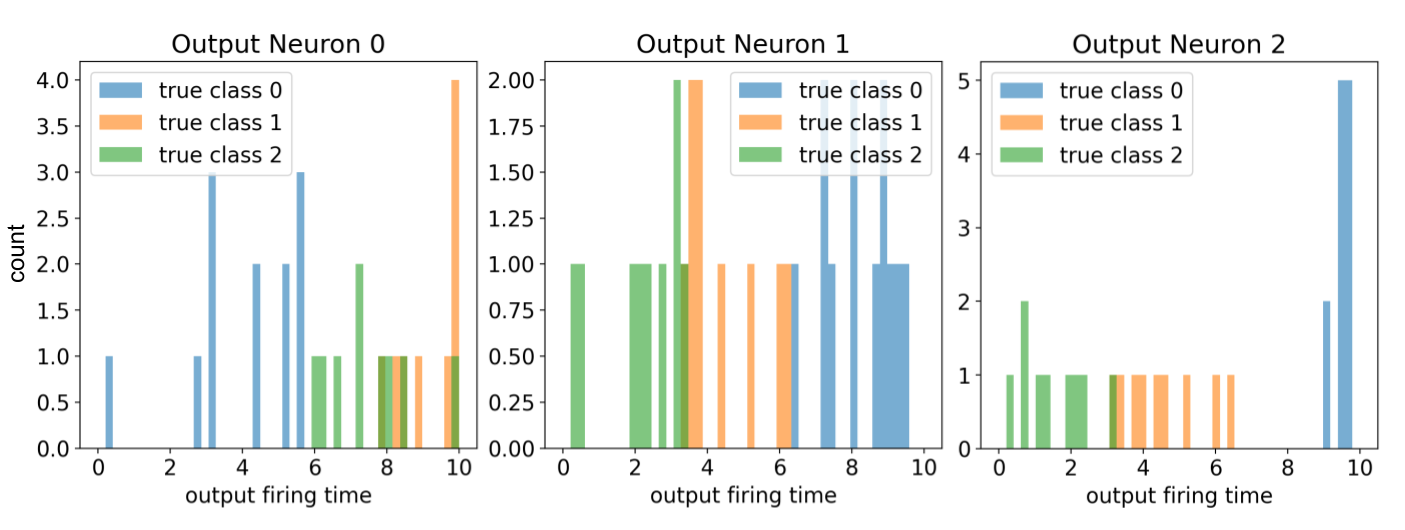}
    \caption{A histogram of the first output firing times of each label neuron, given unseen test data. \textbf{(Top)} At random initialization, firing times look the same across all label neurons. \textbf{(Bottom)} After training, the firing times are clearly separated into the 3 classes, and all test examples belonging to the same class as the corresponding label neuron fires earlier than in the other label neurons.}
    \label{fig:iris}
\end{figure}

We also trained SNN using FP on the Iris dataset to demonstrate learning from data with real-valued features. Note one class is linearly separable from the other 2; the latter are not linearly separable from each other \cite{iris_anderson, iris_fisher}. We encoded the input features with a scheme similar to \cite{mt_spike}, but modified to where each real-valued feature $n_i$ is transformed into a firing time via the transformation $T \cdot (1 - \frac{n_i - \min (n_i)}{\max (n_i) - \max(n_i)})$, where $T$ is the maximum time horizon and the min/max of a feature is taken over the whole dataset. After training a small 4-10-3 network, we achieve 100\% test accuracy (compare to 93.3\% for MT-1 (4-25-1) and 96.7\% for an MLP ANN (4-25-3) in \cite{mt_spike}). Again, the network is able to learn weights to push the true label output neurons to fire earlier than the others, since our loss function is minimized when all the correct label neurons fire before other output neurons. An illustration of this effect is shown in Figure \ref{fig:iris}.

\subsection{Yin-Yang Dataset}
We also implemented FP to train SNN on the Yin-Yang dataset which is a two-dimensional and non-linearly separable dataset \cite{kriener2022yinyang}. The Yin-Yang dataset requires a multi-layer model, as a shallow classifier achieves around 64\% accuracy, thus it requires a hidden layer and backpropagation (or forward-propagation in our case) for gradient-based learning to achieve higher accuracy, as noted also in \cite{eventbasedexactgrad2021}. 

We used a loss based on the earliest spike times of the 3 output neurons, as in \cite{eventbasedexactgrad2021, goltz2021fast} defined as
\begin{align*}
    \mathcal{L} &= - \frac{1}{N_{\textnormal{batch}}} \bigg[ \sum_{i=1}^{N_{\textnormal{batch}}} \log \bigg( \frac{e^{-f_{i, l(i)} / \tau_0}}{ \sum_{j=1}^3 e^{-f_{i, j} / \tau_0} } \bigg) \\
    &~~~~~~~~~~~~~~~~~~~~~~~~~~~~~~~~~~~~~~~~~~~~~~~+ \gamma \big( e^{f_{i, l(i)} / \tau_1} - 1 \big)  \bigg],
\end{align*}
where $f_{i, j}$ is the first spike time of neuron $j$ for the $i^{\textnormal{th}}$ example and $l(i)$ is the index of the correct label for the $i^{\textnormal{th}}$ example. The second term is a regularization term which encourages earlier spike times for the true label neuron, its influence on the total loss controlled by $\gamma$. 

\begin{figure}
\centering
\includegraphics[width=0.4\columnwidth]{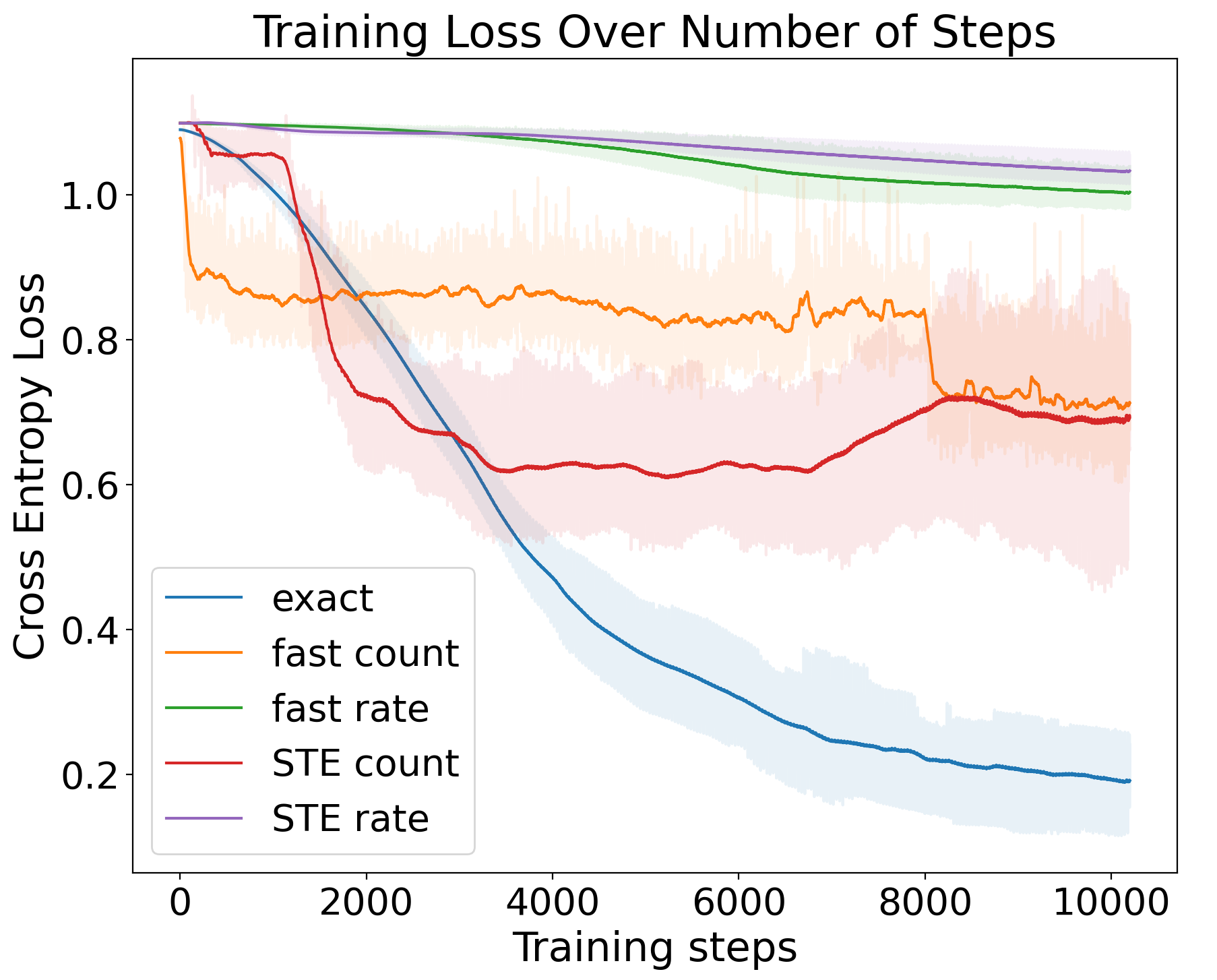}
\includegraphics[width=0.2\columnwidth]{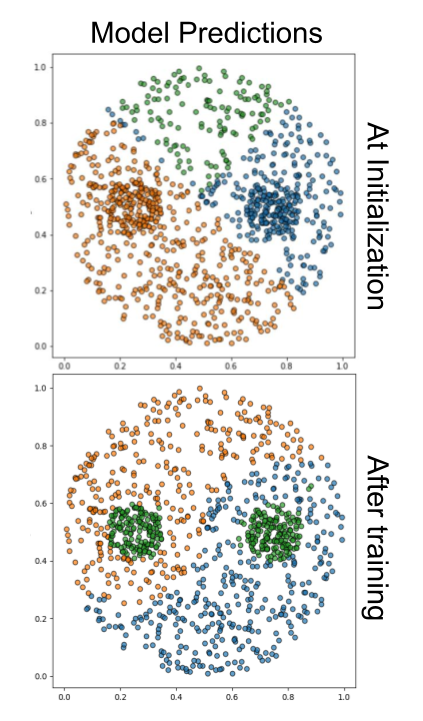}
\caption{\textbf{(Left)} Comparison to surrogate gradients. The plot shows the change in training loss over time for training SNN with exact and surrogate gradients, the fast sigmoid function and straight-through estimator each with a count-based and spike-rate cross entropy loss. \textbf{(Right)} A comparison of model predictions at random initialization, versus after training.}
\label{fig_loss_pred}
\end{figure}

\paragraph{Comparing to surrogate methods.}First, to compare training with surrogate gradient methods, we used the snnTorch library \cite{snntorch} to train equivalent models\footnote{Many surrogate methods are usually not compatible with training using temporal losses, as noted also by \cite{snntorch} that often the first spike time is non-differentiable with respect to the spikes themselves. To fairly compare to surrogate methods, instead we used both a spike count-based cross entropy loss and a spike rate cross entropy loss. The former calculates cross entropy from the number of spikes emitted by output neurons, with the network learning to fire more at the label neuron, 
and the latter accumulates cross entropy loss at each time step, with the network learning to fire continuously at the label neuron and others to be silent.
}, using the same hyperparameters and initializations, but with surrogate gradients. Fig. \ref{fig_loss_pred} (left) compares training with exact gradient (our method) with using the fast sigmoid \cite{zenke_superspike_2018} surrogate function and the straight-through estimator \cite{bengio_ste}, with both count-based cross entropy loss and a spike rate cross entropy loss. (See footnote.) All models at initialization have around 30-36\% accuracy and cross entropy loss around 1.09-1.1, but at the end of 300 epochs of training, using exact gradients results in faster loss reduction (as one might expect).

\paragraph{Evaluation.} After repeating the experiment with 10 random initializations, a 2-layer SNN model trained with FP obtains a test accuracy with mean {\bf 95.0}(0.83)\%, comparable to \cite{goltz2021fast} reporting 95.9(0.7)\%. It is worth noting that training only involved using the exact gradients for SGD, without employing other heuristics in \cite{goltz2021fast}, which include a flat weight bump (increase weights a fixed amount) whenever the proportion of non-spiking neurons is above a certain threshold, among others. These experiments offer a proof of concept that the network is able to learn by using exact gradients. We hope our work will provide a rigorous stepping stone for developing or improving a training library for SNNs.

\section{Discussion}\label{sec:discussion}
Our framework offers an alternative view of the differentiability of SNN w.r.t. network weights and provides a new algorithm, forward-propagation (FP) to calculate gradients of SNN by accumulating information in the forward pass of the network. Our results apply generally to networks with arbitrary recurrent connections, and the ideas can be generalized to other Spike Response Models (SRM). Our gradient method can be used with other algorithms that can simulate the forward pass dynamics, and the FP algorithm dependence on just the causal graph of firing times allows for self-contained formulas which can be often be computed in parallel, e.g., in feed-forward networks. The operations used to compute gradients via FP are also simple and require solving a lower triangular linear system, which can be done quickly.

An interesting by-product of our framework is the fact that our formulas resemble surrogate gradient methods and Hebbian learning. For instance, \cite{zenke_superspike_2018} uses the negative half of the sigmoid function to smooth out the discrete spiking behavior. Our framework captures a natural smoothing exponential kernel already present in the exact version. (See Remark \ref{rem:surrogate_grad}.) Further, the way the smooth kernels $y_{ij}(t)$ between two neurons $i$ and $j$ that appear in the gradient computation resembles Hebbian learning where if there are more spikes from $i$ to $j$ the kernel $y_{ij}(t)$ becomes larger, thus, causing the gradient w.r.t. the connecting weight $W_{ij}$ to become larger. This has a Hebbian flavor where more firing/activation causes the connecting weight $W_{ij}$ to be rewarded (for negative gradient) or punished (for positive gradients) more strongly. These relationships can be of their own interest.

\newpage
\nocite{*}
\bibliographystyle{plain}
\bibliography{refs}

\begin{thebibliography}{10}

\bibitem{true_north_2015}
Filipp Akopyan, Jun Sawada, Andrew Cassidy, Rodrigo Alvarez-Icaza, John Arthur,
  Paul Merolla, Nabil Imam, Yutaka Nakamura, Pallab Datta, Gi-Joon Nam, Brian
  Taba, Michael Beakes, Bernard Brezzo, Jente~B. Kuang, Rajit Manohar,
  William~P. Risk, Bryan Jackson, and Dharmendra~S. Modha.
\newblock Truenorth: Design and tool flow of a 65 mw 1 million neuron
  programmable neurosynaptic chip.
\newblock {\em IEEE Transactions on Computer-Aided Design of Integrated
  Circuits and Systems}, 34(10):1537--1557, 2015.

\bibitem{iris_anderson}
Edgar Anderson.
\newblock The species problem in iris.
\newblock {\em Annals of the Missouri Botanical Garden}, 23(3):457--509, 1936.

\bibitem{bellec_lstm_NEURIPS2018}
Guillaume Bellec, Darjan Salaj, Anand Subramoney, Robert Legenstein, and
  Wolfgang Maass.
\newblock Long short-term memory and learning-to-learn in networks of spiking
  neurons.
\newblock In S.~Bengio, H.~Wallach, H.~Larochelle, K.~Grauman, N.~Cesa-Bianchi,
  and R.~Garnett, editors, {\em Advances in Neural Information Processing
  Systems}, volume~31. Curran Associates, Inc., 2018.

\bibitem{bengio_ste}
Yoshua Bengio, Nicholas Léonard, and Aaron Courville.
\newblock Estimating or propagating gradients through stochastic neurons for
  conditional computation, 2013.

\bibitem{spike_prop}
Sander~M. Boht{\'e}, Joost~N. Kok, and Han~La Poutr{\'e}.
\newblock Spikeprop: backpropagation for networks of spiking neurons.
\newblock In {\em ESANN}, 2000.

\bibitem{burkitt_integrate_fire_2006}
Anthony Burkitt.
\newblock A review of the integrate-and-fire neuron model: I. homogeneous
  synaptic input.
\newblock {\em Biological cybernetics}, 95:1--19, 08 2006.

\bibitem{cao_snn_object_recognition_2015}
Yongqiang Cao, Yang Chen, and Deepak Khosla.
\newblock Spiking deep convolutional neural networks for energy-efficient
  object recognition.
\newblock {\em International Journal of Computer Vision}, 113:54--66, 05 2015.

\bibitem{hebbian_overview_2013}
Yoonsuck Choe.
\newblock {\em Hebbian Learning}, pages 1--5.
\newblock Springer New York, New York, NY, 2013.

\bibitem{comsa_temporal_coding_2020}
Iulia Comsa, Thomas Fischbacher, Krzysztof Potempa, Andrea Gesmundo, Luca
  Versari, and Jyrki Alakuijala.
\newblock Temporal coding in spiking neural networks with alpha synaptic
  function.
\newblock pages 8529--8533, 05 2020.

\bibitem{loihi_2018}
Mike Davies, Narayan Srinivasa, Tsung-Han Lin, Gautham Chinya, Yongqiang Cao,
  Sri~Harsha Choday, Georgios Dimou, Prasad Joshi, Nabil Imam, Shweta Jain,
  Yuyun Liao, Chit-Kwan Lin, Andrew Lines, Ruokun Liu, Deepak Mathaikutty,
  Steven McCoy, Arnab Paul, Jonathan Tse, Guruguhanathan Venkataramanan,
  Yi-Hsin Weng, Andreas Wild, Yoonseok Yang, and Hong Wang.
\newblock Loihi: A neuromorphic manycore processor with on-chip learning.
\newblock {\em IEEE Micro}, 38(1):82--99, 2018.

\bibitem{deng2012mnist}
Li~Deng.
\newblock The mnist database of handwritten digit images for machine learning
  research.
\newblock {\em IEEE Signal Processing Magazine}, 29(6):141--142, 2012.

\bibitem{diehl_unsupervised_snn_2015}
Peter Diehl and Matthew eCook.
\newblock Unsupervised learning of digit recognition using
  spike-timing-dependent plasticity.
\newblock {\em Frontiers in Computational Neuroscience}, 9, 08 2015.

\bibitem{Ding2021OptimalAC}
Jianhao Ding, Zhaofei Yu, Yonghong Tian, and Tiejun Huang.
\newblock Optimal ann-snn conversion for fast and accurate inference in deep
  spiking neural networks.
\newblock {\em ArXiv}, abs/2105.11654, 2021.

\bibitem{snntorch}
Jason~K Eshraghian, Max Ward, Emre Neftci, Xinxin Wang, Gregor Lenz, Girish
  Dwivedi, Mohammed Bennamoun, Doo~Seok Jeong, and Wei~D Lu.
\newblock Training spiking neural networks using lessons from deep learning.
\newblock {\em arXiv preprint arXiv:2109.12894}, 2021.

\bibitem{eshraghian2021training_snntorch}
Jason~K Eshraghian, Max Ward, Emre Neftci, Xinxin Wang, Gregor Lenz, Girish
  Dwivedi, Mohammed Bennamoun, Doo~Seok Jeong, and Wei~D Lu.
\newblock Training spiking neural networks using lessons from deep learning.
\newblock {\em arXiv preprint arXiv:2109.12894}, 2021.

\bibitem{esser_conv_2016}
Steven~K. Esser, Paul~A. Merolla, John~V. Arthur, Andrew~S. Cassidy,
  Rathinakumar Appuswamy, Alexander Andreopoulos, David~J. Berg, Jeffrey~L.
  McKinstry, Timothy Melano, Davis~R. Barch, Carmelo di~Nolfo, Pallab Datta,
  Arnon Amir, Brian Taba, Myron~D. Flickner, and Dharmendra~S. Modha.
\newblock Convolutional networks for fast, energy-efficient neuromorphic
  computing.
\newblock {\em Proceedings of the National Academy of Sciences},
  113(41):11441--11446, 2016.

\bibitem{iris_fisher}
R.~A. Fisher.
\newblock The use of multiple measurements in taxonomic problems.
\newblock {\em Annals of Eugenics}, 7(2):179--188, 1936.

\bibitem{srm}
Wulfram Gerstner.
\newblock Time structure of the activity in neural network models.
\newblock {\em Phys. Rev. E}, 51:738--758, Jan 1995.

\bibitem{gerstner_kistler_naud_paninski_2014}
Wulfram Gerstner, Werner~M. Kistler, Richard Naud, and Liam Paninski.
\newblock {\em Neuronal Dynamics: From Single Neurons to Networks and Models of
  Cognition}.
\newblock Cambridge University Press, 2014.

\bibitem{dendritic}
Albert Gidon, Timothy~Adam Zolnik, Pawel Fidzinski, Felix Bolduan, Athanasia
  Papoutsi, Panayiota Poirazi, Martin Holtkamp, Imre Vida, and Matthew~Evan
  Larkum.
\newblock Dendritic action potentials and computation in human layer 2/3
  cortical neurons.
\newblock {\em Science}, 367(6473):83--87, 2020.

\bibitem{goltz2021fast}
Julian Göltz, Laura Kriener, Andreas Baumbach, Sebastian Billaudelle, Oliver
  Breitwieser, Benjamin Cramer, Dominik Dold, Akos~Ferenc Kungl, Walter Senn,
  Johannes Schemmel, Karlheinz Meier, and Mihai~Alexandru Petrovici.
\newblock Fast and energy-efficient neuromorphic deep learning with first-spike
  times, 2021.

\bibitem{ho_conversion_2021}
Nguyen-Dong Ho and Ik-Joon Chang.
\newblock Tcl: an ann-to-snn conversion with trainable clipping layers.
\newblock In {\em 2021 58th ACM/IEEE Design Automation Conference (DAC)}, pages
  793--798, 2021.

\bibitem{huh_gd_NEURIPS2018}
Dongsung Huh and Terrence~J Sejnowski.
\newblock Gradient descent for spiking neural networks.
\newblock In S.~Bengio, H.~Wallach, H.~Larochelle, K.~Grauman, N.~Cesa-Bianchi,
  and R.~Garnett, editors, {\em Advances in Neural Information Processing
  Systems}, volume~31. Curran Associates, Inc., 2018.

\bibitem{jeffares2022spikeinspired}
Alan Jeffares, Qinghai Guo, Pontus Stenetorp, and Timoleon Moraitis.
\newblock Spike-inspired rank coding for fast and accurate recurrent neural
  networks.
\newblock In {\em International Conference on Learning Representations}, 2022.

\bibitem{macro_micro_backprop}
Yingyezhe Jin, Wenrui Zhang, and Peng Li.
\newblock Hybrid macro/micro level backpropagation for training deep spiking
  neural networks.
\newblock In S.~Bengio, H.~Wallach, H.~Larochelle, K.~Grauman, N.~Cesa-Bianchi,
  and R.~Garnett, editors, {\em Advances in Neural Information Processing
  Systems}, volume~31. Curran Associates, Inc., 2018.

\bibitem{kempter_hebbian_1999}
Richard Kempter, Wulfram Gerstner, and Leo van Hemmen.
\newblock Hebbian learning and spiking neurons.
\newblock {\em Phys. Rev. E}, 59, 04 1999.

\bibitem{kornijcuk_lif_2016}
Vladimir Kornijcuk, Hyungkwang Lim, Jun~Yeong Seok, Guhyun Kim, Seong~Keun Kim,
  Inho Kim, Byung~Joon Choi, and Doo~Seok Jeong.
\newblock Leaky integrate-and-fire neuron circuit based on floating-gate
  integrator.
\newblock {\em Frontiers in Neuroscience}, 10, 2016.

\bibitem{kriener2022yinyang}
Laura Kriener, Julian Göltz, and Mihai~A. Petrovici.
\newblock The yin-yang dataset, 2022.

\bibitem{kuroe-synthesizing_2010}
Y.~Kuroe and H.~Iima.
\newblock A learning method for synthesizing spiking neural oscillators.
\newblock In {\em The 2006 IEEE International Joint Conference on Neural
  Network Proceedings}, pages 3882--3886, 2006.

\bibitem{kuroe_adjoint_2010}
Yasuaki Kuroe and Tomokazu Ueyama.
\newblock Learning methods of recurrent spiking neural networks based on
  adjoint equations approach.
\newblock In {\em The 2010 International Joint Conference on Neural Networks
  (IJCNN)}, pages 1--8, 2010.

\bibitem{lee_stdp_pretrain_2018}
Chankyu Lee, Priyadarshini Panda, Gopalakrishnan Srinivasan, and Kaushik Roy.
\newblock Training deep spiking convolutional neural networks with stdp-based
  unsupervised pre-training followed by supervised fine-tuning.
\newblock {\em Frontiers in Neuroscience}, 12, 2018.

\bibitem{mt_spike}
Tao Liu, Zihao Liu, Fuhong Lin, Yier Jin, Gang Quan, and Wujie Wen.
\newblock Mt-spike: A multilayer time-based spiking neuromorphic architecture
  with temporal error backpropagation.
\newblock pages 450--457, 11 2017.

\bibitem{lobov_stdp_2020}
Sergey~A. Lobov, Alexey~N. Mikhaylov, Maxim Shamshin, Valeri~A. Makarov, and
  Victor~B. Kazantsev.
\newblock Spatial properties of stdp in a self-learning spiking neural network
  enable controlling a mobile robot.
\newblock {\em Frontiers in Neuroscience}, 14, 2020.

\bibitem{maass_third_gen}
Wolfgang Maass.
\newblock Networks of spiking neurons: The third generation of neural network
  models.
\newblock {\em Neural Networks}, 10(9):1659--1671, 1997.

\bibitem{plasticity}
Timoleon Moraitis, Abu Sebastian, and Evangelos Eleftheriou.
\newblock Optimality of short-term synaptic plasticity in modelling certain
  dynamic environments, 2020.

\bibitem{temporal_coding}
Hesham Mostafa.
\newblock Supervised learning based on temporal coding in spiking neural
  networks.
\newblock {\em IEEE Transactions on Neural Networks and Learning Systems}, PP,
  06 2016.

\bibitem{Neftci2019SurrogateGL}
Emre~O. Neftci, Hesham Mostafa, and Friedemann Zenke.
\newblock Surrogate gradient learning in spiking neural networks.
\newblock {\em ArXiv}, abs/1901.09948, 2019.

\bibitem{nmnist}
Garrick Orchard, Ajinkya Jayawant, Gregory~K. Cohen, and Nitish Thakor.
\newblock Converting static image datasets to spiking neuromorphic datasets
  using saccades.
\newblock {\em Frontiers in Neuroscience}, 9, 2015.

\bibitem{panda_snn_residual_2020}
Priyadarshini Panda, Aparna Aketi, and Kaushik Roy.
\newblock Toward scalable, efficient, and accurate deep spiking neural networks
  with backward residual connections, stochastic softmax, and hybridization.
\newblock {\em Frontiers in Neuroscience}, 14:653, 06 2020.

\bibitem{pfeiffer_opportunities_2018}
Michael Pfeiffer and Thomas Pfeil.
\newblock Deep learning with spiking neurons: Opportunities and challenges.
\newblock {\em Frontiers in Neuroscience}, 12, 2018.

\bibitem{compneurobook}
Patrick~D. Roberts.
\newblock {\em Synaptic Dynamics: Overview}, pages 1--4.
\newblock Springer New York, 2013.

\bibitem{roy_snn_computing_2019}
Kaushik Roy, Akhilesh Jaiswal, and Priyadarshini Panda.
\newblock Towards spike-based machine intelligence with neuromorphic computing.
\newblock {\em Nature}, 575:607--617, 11 2019.

\bibitem{rueckauer_conversion_2017}
Bodo Rueckauer, Iulia-Alexandra Lungu, Yuhuang Hu, Michael Pfeiffer, and
  Shih-Chii Liu.
\newblock Conversion of continuous-valued deep networks to efficient
  event-driven networks for image classification.
\newblock {\em Frontiers in Neuroscience}, 11, 2017.

\bibitem{ruf_hebbbian_2006}
Berthold Ruf and Michael Schmitt.
\newblock {\em Hebbian learning in networks of spiking neurons using temporal
  coding}, pages 380--389.
\newblock 04 2006.

\bibitem{safa_convsnn_2021}
Ali Safa, Francky Catthoor, and Georges~G.E. Gielen.
\newblock Convsnn: A surrogate gradient spiking neural framework for radar
  gesture recognition.
\newblock {\em Software Impacts}, 10:100131, 2021.

\bibitem{Selvaratnam2000LearningMO}
Kukan Selvaratnam, Yasuaki Kuroe, and Takehiro Mori.
\newblock Learning methods of recurrent spiking neural networks.
\newblock 2000.

\bibitem{shrestha_slayer_NEURIPS2018}
Sumit~Bam Shrestha and Garrick Orchard.
\newblock Slayer: Spike layer error reassignment in time.
\newblock In S.~Bengio, H.~Wallach, H.~Larochelle, K.~Grauman, N.~Cesa-Bianchi,
  and R.~Garnett, editors, {\em Advances in Neural Information Processing
  Systems}, volume~31. Curran Associates, Inc., 2018.

\bibitem{tavanaei_overview_2018}
Amirhossein Tavanaei, Masoud Ghodrati, Saeed~Reza Kheradpisheh, Timothée
  Masquelier, and Anthony Maida.
\newblock Deep learning in spiking neural networks.
\newblock {\em Neural Networks}, 04 2018.

\bibitem{wang_review_2020}
Xiangwen Wang, Xianghong Lin, and Xiaochao Dang.
\newblock Supervised learning in spiking neural networks: A review of
  algorithms and evaluations.
\newblock {\em Neural Networks}, 125:258--280, 05 2020.

\bibitem{eventbasedexactgrad2021}
Timo Wunderlich and Christian Pehle.
\newblock Event-based backpropagation can compute exact gradients for spiking
  neural networks.
\newblock {\em Scientific Reports}, 11:12829, 06 2021.

\bibitem{zenke_superspike_2018}
Friedemann Zenke and Surya Ganguli.
\newblock {SuperSpike: Supervised Learning in Multilayer Spiking Neural
  Networks}.
\newblock {\em Neural Computation}, 30(6):1514--1541, 06 2018.

\bibitem{spike_training_rsnn}
Wenrui Zhang and Peng Li.
\newblock Spike-train level backpropagation for training deep recurrent spiking
  neural networks.
\newblock In H.~Wallach, H.~Larochelle, A.~Beygelzimer, F.~d\textquotesingle
  Alch\'{e}-Buc, E.~Fox, and R.~Garnett, editors, {\em Advances in Neural
  Information Processing Systems}, volume~32. Curran Associates, Inc., 2019.

\end{thebibliography}

\newpage
\section{Appendix}

\subsection{Alternative Proof of Theorem \ref{thm:response}}\label{appendix:alt_proof_thm3}

{\bf Proof (ii)}: Here we provide a more rigorous proof based on induction on the number of firing times $F_i:=|\mathcal{F}_i|$ of the neuron $i$. 

We first check the base of the induction. If there are no firing times, i.e., $\mathcal{F}_i= \emptyset$ and $F_i=0$, then there is no source of non-linearity and the neuron is a fully linear system. Thus, the response of the neuron to the input weighted synaptic current $I^s_i(t)$ is given, as in \eqref{eq:snn_lin_dyn}, by
$$V_i(t)= \sum_{j\in \mathcal{N}_i} W_{ji} y_{ji}(t),$$
which yields the desired result since, for $\mathcal{F}_i=\emptyset$, the second term $-\sum_{f \in \mathcal{F}_i} \theta_i h^n_{i}(t-f)$ is zero. This confirms the base of induction for $F_i=0$.

Now let us assume that $\mathcal{F}_i \not = \emptyset$ and the neuron $i$ has fired at least once ($F_i\geq 1$).
Here, we can still check that result holds for all time $t \in [ 0, f_1 )$ before the first firing time $f_1$ because before the first firing time the circuit is completely linear (thus, the first term) and the second term is equal to zero as $h^n_{i}(t-f_1)=e^{\beta_{i}(t-f_1)}u(t-f_1)$ is equal to zero for all $t<f_1$ (due to causality and the fact that $u(t-f_1)=0$ for $t<f_1$).

Now we prove that if the result is true for $t \in [0, f^{k})$ it remains true for $t \in [f_k, f_{k+1})$ where we denote the $k$-th and $(k+1)$-th firing times by $f_k$ and $f_{k+1}$ and apply the convention that $f_k=\infty$ for $k>F_i$.

To prove this, we first note that the weighted synaptic current (see, e.g., Fig.\,\ref{fig:delta_circuit_model}) coming from the neurons $\mathcal{N}_i$ is given by 
$$I^s_i(t)= \sum_{j \in \mathcal{N}_i} W_{ji} \sum_{g \in \mathcal{F}_i} h^s_{j}(t-g)$$
for all times $t\geq 0$. Also, note that since synapses are always linear, this is true independent of whether there is any firing and potential drop at the neuron $i$. At the firing time $f_k$ the value of potential drops to 
$V_i^{(k)} = V_i(f_k) - \theta_i$. 
Thus, to prove the result, we need to find and verify the response of the neuron to the synaptic current $I^s_i(t)$ for $t\in [f_k, f_{k+1})$ starting from the initial value $V_i^{(k)}$. Here again we note that starting from $f_k$ the system is again linear until the next firing time $f_{k+1}$. Thus, we can again apply the super position principle for linear systems to decompose the response into two parts: (a)\,response to the initial condition $V_i^{(k)}$ and (b)\,response to the input synaptic current $I^s_i(t)$. 

From the linearity and time-invariance of RC circuits, (a) is simply given by 
\begin{align*}
    V_i^{(a)}(t)&= V_i^{(k)} h^n_{i}(t-f_k) \\
    &= V_i^{(k)} e^{-\beta_{i} (t-f_k)}u(t-f_k)\\
    &= V_i(f_k) e^{-\beta_{i} (t-f_k)}u(t-f_k) - \theta_i h^n_{i}(t-f_k),
\end{align*}
where $h^n_{i}(t)=e^{-\beta_{i} t} u(t)$ is the impulse response of the neuron $i$.

The response to the synaptic current in the time interval $t \in [f_k, f_{k+1})$ is also given by 
\begin{align*}
    V_i^{(b)}(t) &\stackrel{(i)}{=} I^s_i(t)u(t-f_k) \star h^n_{i}(t) \\
    &= \int_0^\infty I^s_i(\lambda) u(\lambda -f_k)  h^n_{i}(t- \lambda) d \lambda\\
    &\stackrel{(ii)}{=} \int_{f_k}^t  I^s_i(\lambda) h^n_{i}(t- \lambda) d \lambda\\
    &= \int_0^t  I^s_i(\lambda) h^n_{i}(t- \lambda) d \lambda - \int_0^{f_k}  I^s_i(\lambda) h^n_{i}(t- \lambda) d \lambda\\
    &= I^s_i(t)\star h^n_{i}(t) - \int_0^{f_k}  I^s_i(\lambda) e^{-\beta_{i}(t- \lambda)} d \lambda\\
    &= I^s_i(t)\star h^n_{i}(t) - e^{-\beta_{i}(t- f_k)} \int_0^{f_k}  I^s_i(\lambda) e^{-\beta_{i}(f_k- \lambda)} d \lambda\\
    &= I^s_i(t)\star h^n_{i}(t)  - I^s_i(t)\star h^n_{i}(t) \Big | _{t=f_k} \times e^{-\beta_{i}(t- f_k)}, \label{eq_induction}
\end{align*}
where in $(i)$ we multiplied $I^s_i(t)$ with $u(t-f_k)$ to remove the effect of the synaptic current before $f_k$ (since, due to causality, it cannot affect the neuron potential in the time interval $t \in [ f_k, f_{k+1} )$), where in $(ii)$ we used the fact that, due to causality, $h_{ni}(t-\lambda)=0$ for $\lambda>t$, and that $u(\lambda -f_k)$ is zero for $\lambda < f_k$. 

From the induction hypothesis applied to $f_k \in [0, f_k]$, we have that 
\begin{align*}
    V_i(f_k)&= I^s_i(t)\star h^n_{i}(t) \Big |_{t=f_k} - \theta_i \sum_{l=1}^{k-1} h^n_{i}(f_k - f_l)\\
    &= I^s_i(t)\star h^n_{i}(t) \Big |_{t=f_k} - \theta_i \sum_{l=1}^{k-1} h^n_{i}(f_k - f_l)\\
    &= I^s_i(t)\star h^n_{i}(t) \Big |_{t=f_k} - \theta_i \sum_{l=1}^{k-1} e^{-\beta_{i}(f_k - f_l)} \\
    &= I^s_i(t)\star h^n_{i}(t) \Big |_{t=f_k} - \theta_i e^{\beta_{i}(t-f_k)} \sum_{l=1}^{k-1} e^{-\beta_{i}(t - f_l)}\\
    &= I^s_i(t)\star h^n_{i}(t) \Big |_{t=f_k} - \theta_i e^{\beta_{i}(t-f_k)} \sum_{l=1}^{k-1} h^n_{i}(t-f_l).
\end{align*}
Therefore, after simplification, we obtain that
\begin{align}
    I^s_i(t)&\star h^n_{i}(t) \Big | _{t=f_k} \times e^{-\beta_{i}(t- f_k)} \\
    &= V_i(f_k) e^{-\beta_{i}(t-f_k)} + \theta_i \sum_{l=1}^{k-1} h^n_{i}(t-f_l).
\end{align}
Replacing in \eqref{eq_induction}, therefore, we obtain
\begin{align}
    V_i^{(b)}(t)&= I^s_i(t)\star h^n_{i}(t) \\
    &- V_i(f_k) e^{-\beta_{i}(t-f_k)} - \theta_i \sum_{l=1}^{k-1} h^n_{i}(t-f_l).
\end{align}
Applying the super position principle, we have
\begin{align*}
    V_i(t)&=V_i^{(a)}(t) + V_i^{(b)}(t)\\
    &= I^s_i(t)\star h^n_{i}(t) - \theta_i \sum_{l=1}^{k-1} h^n_{i}(t-f_l) - \theta_i h^n_{i}(t-f_k) \\
    &= I^s_i(t)\star h^n_{i}(t) - \theta_i \sum_{l=1}^{k} h^n_{i}(t-f_l)\\
    &= \Big ( \sum_{j \in \mathcal{N}_i} W_{ji} \sum_{g \in \mathcal{F}_i} h^s_{j}(t-g) \Big ) \star h^n_{i}(t) - \theta_i \sum_{l=1}^{k} h^n_{i}(t-f_l)\\
    &=  \sum_{j \in \mathcal{N}_i} W_{ji} \sum_{g \in \mathcal{F}_i} h_{ji}(t-g) - \theta_i \sum_{l=1}^{k} h^n_{i}(t-f_l)\\
    &= \sum_{j \in \mathcal{N}_i} W_{ji} y_{ji}(t) - \theta_i \sum_{f \in \mathcal{F}_i}  h^n_{i}(t-f),
\end{align*}
where in the last equation we used the fact that $h^n_{i}(t-f)=0$ for $t \in [f_k, f_{k+1})$ and for $f>f_{k+1}$. This validates the result for $t \in [f_k, f_{k+1})$, and verifies the induction. This completes the proof.

\subsection{Implicit Function Theorem}\label{appendix:ift}
In many problem in machine learning, statistics, control theory, mathematics, etc. we use a collection of variables to track/specify the state of an algorithm, a dynamical system, etc. However, in practice, these variables are not completely free and are connected to each other via specific constraints. In such cases, we are always interested to know the functional relation between these variables, namely, how changing some variables affect the others (sensitivity analysis). IFT theorem provides a rigorous method for these types of analyses when the variables are connected through differentiable equality constraints, as illustrated in the following theorem.
\begin{theorem}[Implicit Function Theorem]
Let $\phi: \mathbb{R}^n \times \mathbb{R}^m \to \mathbb{R}^m$ be a differentiable function and let $\mathcal{Z}=\{(x,y) \in \mathbb{R}^n \times \mathbb{R}^m: \phi(x,y)=0\}$ be the zero-set of $\phi$. Suppose that $\mathcal{Z} \not = \emptyset$ and let $(x_0,y_0) \in \mathcal{Z}$ be an arbitrary point. Also, let $\frac{\partial \phi}{\partial y} \phi(x_0,y_0)$ be the $m\times m$ matrix of partial derivatives w.r.t. $y$ and assume that it is non-singular, i.e., $\det\big ( \frac{\partial \phi}{\partial y} (x_0,y_0) \big ) \not = 0$. Then, 
\begin{itemize}
    \item There is an open neighborhood $\mathcal{N}_x$ around $x_0$ and an open neighborhood $\mathcal{N}_\mathrm{y}$ around $y_0$ such that $\frac{\partial \phi}{\partial y} \phi (x,y)$ is non-singular for all $(x,y) \in \mathcal{N}:=\mathcal{N}_\mathrm{x} \times \mathcal{N}_\mathrm{y}$ (including of course the original $(x_0,y_0)$.
    
    \item There is a function $\psi: \mathcal{N}_\mathrm{x} \to \mathcal{N}_\mathrm{y}$ such that $(x,\psi(x))$ belongs to the zero set $\mathcal{Z}$, namely, $\phi(x,\psi(x))=0$, for all $x \in \mathcal{N}_\mathrm{x}$; therefore, the variables $y$ in $\mathcal{N}_\mathrm{y}$ can be written as a function $y=\psi(x)$ of the variables $x$ in $\mathcal{N}_\mathrm{x}$.
    
    \item $\psi$ is a differentiable function of $x$ for $x \in \mathcal{N}_\mathrm{x}$ and 
    \begin{align}
        \frac{\partial \phi}{\partial y}  \times \frac{\partial \psi}{\partial x} + \frac{\partial \phi}{\partial x}=0,
    \end{align}
    which from the non-singularity of $\frac{\partial \phi}{\partial y}$ yields
    \begin{align}
         \frac{\partial \psi}{\partial x} = - \Big ( \frac{\partial \phi}{\partial y} \Big ) ^ {-1} \times \frac{\partial \psi}{\partial x}.
    \end{align}
    
\end{itemize}

\end{theorem}

\begin{figure}[ht]
\centering
\includegraphics[width=0.6\columnwidth]{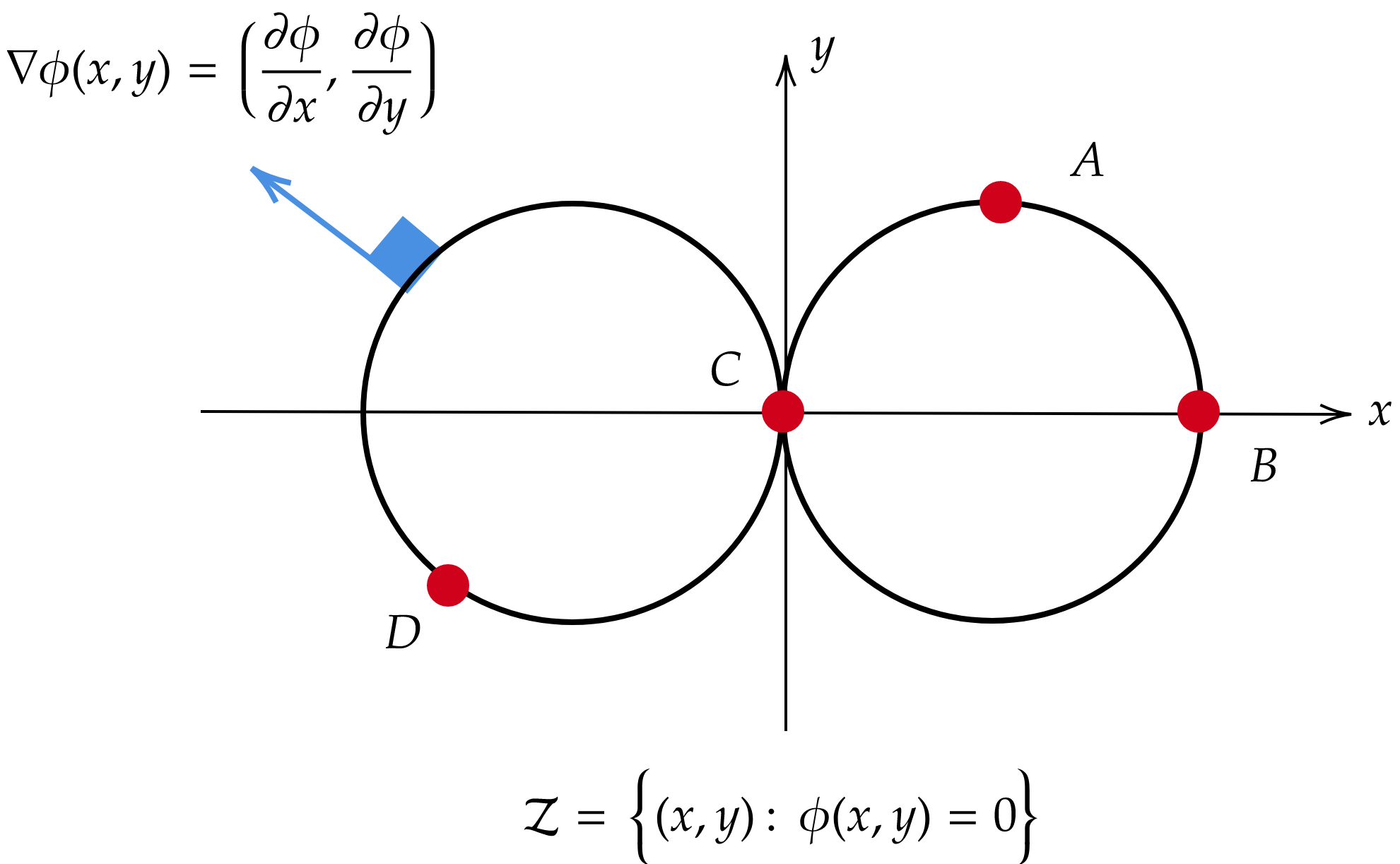}
\caption{Illustration of the implicit function theorem.}
\label{fig_ift}
\end{figure}

{\bf Example 1.} Fig.\,\ref{fig_ift} illustrates the zero-set $\mathcal{Z}=\{(x,y): \phi(x,y)=0\}$ of a function $\phi: \mathbb{R}^2 \to \mathbb{R}$. To investigate the conditions of the implicit function theorem, we first note that the gradient of $\phi$ denoted by $\nabla \phi=(\frac{\partial \phi}{\partial x}, \frac{\partial \phi}{\partial y})$  is always orthogonal to the level-set (here the zero-set) of $\phi$. Thus, by observing the orthogonal vector to curve, we can verify if $\frac{\partial \phi}{\partial x}$ or $\frac{\partial \phi}{\partial y}$ are non-singular (non-zero in the scalar case we consider here).
We investigate several cases:
\begin{itemize}
    \item Point $C$: gradient vector does not exist, so the assumptions of the IFT are not fulfilled. One can also see that at $C$ one cannot write neither $x$ as a function of $y$ nor $y$ as a function of $x$.
    
    \item Point $A$: gradient vector has zero horizontal and non-zero vertical component , i.e., $\frac{\partial \phi}{\partial x} = 0$ and $\frac{\partial \phi}{\partial y} \not = 0$. Thus, from IFT, in a local neighborhood of $A$, one should be able to write only $y$ as a differentiable function of $x$.
    
    \item Point $B$: gradient has zero horizontal component. And, only $x$ can be written as differentiable function of $y$. 
    
    \item Point $D$: gradient has non-zero horizontal and vertical components. So, in a local neighborhood of $D$, one may write both $x$ and $y$ as a differentiable function of the another. 
\end{itemize}

\subsection{Example: Causality and Differentiability}\label{appendix:causal_ex}
In order to track the effects of previous layers' firing times on a current neuron $i$, we can map which firing times of a previous neuron cause the firing of a neuron that it feeds into, and so on through the network. Consider the following simple example of a simple 3 neuron feed-forward network with 1 input dimension:

\tikzset{every picture/.style={line width=0.75pt}} 
\begin{center}
\begin{tikzpicture}[x=0.5pt,y=0.5pt,yscale=-1,xscale=1]

\draw   (157,144) .. controls (157,130.19) and (168.19,119) .. (182,119) .. controls (195.81,119) and (207,130.19) .. (207,144) .. controls (207,157.81) and (195.81,169) .. (182,169) .. controls (168.19,169) and (157,157.81) .. (157,144) -- cycle ;
\draw    (207,144) -- (352.5,143.08) ;
\draw [shift={(354.5,143.06)}, rotate = 179.64] [color={rgb, 255:red, 0; green, 0; blue, 0 }  ][line width=0.75]    (10.93,-3.29) .. controls (6.95,-1.4) and (3.31,-0.3) .. (0,0) .. controls (3.31,0.3) and (6.95,1.4) .. (10.93,3.29)   ;
\draw   (354.5,143.06) .. controls (354.5,129.26) and (365.69,118.06) .. (379.5,118.06) .. controls (393.31,118.06) and (404.5,129.26) .. (404.5,143.06) .. controls (404.5,156.87) and (393.31,168.06) .. (379.5,168.06) .. controls (365.69,168.06) and (354.5,156.87) .. (354.5,143.06) -- cycle ;
\draw   (546,143) .. controls (546,129.19) and (557.19,118) .. (571,118) .. controls (584.81,118) and (596,129.19) .. (596,143) .. controls (596,156.81) and (584.81,168) .. (571,168) .. controls (557.19,168) and (546,156.81) .. (546,143) -- cycle ;
\draw    (404.5,143.06) -- (544,143) ;
\draw [shift={(546,143)}, rotate = 179.97] [color={rgb, 255:red, 0; green, 0; blue, 0 }  ][line width=0.75]    (10.93,-3.29) .. controls (6.95,-1.4) and (3.31,-0.3) .. (0,0) .. controls (3.31,0.3) and (6.95,1.4) .. (10.93,3.29)   ;
\draw    (54.5,94.06) -- (226.5,94.06) ;
\draw    (37.5,144.06) -- (155,144) ;
\draw [shift={(157,144)}, rotate = 179.97] [color={rgb, 255:red, 0; green, 0; blue, 0 }  ][line width=0.75]    (10.93,-3.29) .. controls (6.95,-1.4) and (3.31,-0.3) .. (0,0) .. controls (3.31,0.3) and (6.95,1.4) .. (10.93,3.29)   ;
\draw    (143.5,94.06) -- (143.5,41.06) ;
\draw [shift={(143.5,39.06)}, rotate = 90] [color={rgb, 255:red, 0; green, 0; blue, 0 }  ][line width=0.75]    (10.93,-3.29) .. controls (6.95,-1.4) and (3.31,-0.3) .. (0,0) .. controls (3.31,0.3) and (6.95,1.4) .. (10.93,3.29)   ;
\draw    (200.5,94.06) -- (200.5,41.06) ;
\draw [shift={(200.5,39.06)}, rotate = 90] [color={rgb, 255:red, 0; green, 0; blue, 0 }  ][line width=0.75]    (10.93,-3.29) .. controls (6.95,-1.4) and (3.31,-0.3) .. (0,0) .. controls (3.31,0.3) and (6.95,1.4) .. (10.93,3.29)   ;
\draw    (85.5,94.06) -- (85.5,41.06) ;
\draw [shift={(85.5,39.06)}, rotate = 90] [color={rgb, 255:red, 0; green, 0; blue, 0 }  ][line width=0.75]    (10.93,-3.29) .. controls (6.95,-1.4) and (3.31,-0.3) .. (0,0) .. controls (3.31,0.3) and (6.95,1.4) .. (10.93,3.29)   ;
\draw    (257.5,243.06) -- (429.5,243.06) ;
\draw    (346.5,243.06) -- (346.5,190.06) ;
\draw [shift={(346.5,188.06)}, rotate = 90] [color={rgb, 255:red, 0; green, 0; blue, 0 }  ][line width=0.75]    (10.93,-3.29) .. controls (6.95,-1.4) and (3.31,-0.3) .. (0,0) .. controls (3.31,0.3) and (6.95,1.4) .. (10.93,3.29)   ;
\draw    (403.5,243.06) -- (403.5,190.06) ;
\draw [shift={(403.5,188.06)}, rotate = 90] [color={rgb, 255:red, 0; green, 0; blue, 0 }  ][line width=0.75]    (10.93,-3.29) .. controls (6.95,-1.4) and (3.31,-0.3) .. (0,0) .. controls (3.31,0.3) and (6.95,1.4) .. (10.93,3.29)   ;
\draw    (463.5,86.06) -- (635.5,86.06) ;
\draw    (609.5,86.06) -- (609.5,52.06) -- (609.5,33.06) ;
\draw [shift={(609.5,31.06)}, rotate = 90] [color={rgb, 255:red, 0; green, 0; blue, 0 }  ][line width=0.75]    (10.93,-3.29) .. controls (6.95,-1.4) and (3.31,-0.3) .. (0,0) .. controls (3.31,0.3) and (6.95,1.4) .. (10.93,3.29)   ;

\draw (176,135.4) node [anchor=north west][inner sep=0.75pt]    {$1$};
\draw (373,135.4) node [anchor=north west][inner sep=0.75pt]    {$2$};
\draw (565,134.4) node [anchor=north west][inner sep=0.75pt]    {$3$};
\draw (90.25,123.43) node [anchor=north west][inner sep=0.75pt]    {$w_{1}$};
\draw (273.75,121.93) node [anchor=north west][inner sep=0.75pt]    {$w_{2}$};
\draw (468,120.4) node [anchor=north west][inner sep=0.75pt]    {$w_{3}$};
\draw (55,152.4) node [anchor=north west][inner sep=0.75pt]    {$y_{1}( t)=x( t)$};
\draw (263,151.4) node [anchor=north west][inner sep=0.75pt]    {$y_{2}( \cdot )$};
\draw (458,148.4) node [anchor=north west][inner sep=0.75pt]    {$y_{3}( \cdot )$};
\draw (55,26.4) node [anchor=north west][inner sep=0.75pt]    {$f_{1}^{1}$};
\draw (113,28.4) node [anchor=north west][inner sep=0.75pt]    {$f_{2}^{1}$};
\draw (173,28.4) node [anchor=north west][inner sep=0.75pt]    {$f_{3}^{1}$};
\draw (312,187.4) node [anchor=north west][inner sep=0.75pt]    {$f_{1}^{2}$};
\draw (373,187.4) node [anchor=north west][inner sep=0.75pt]    {$f_{2}^{2}$};
\draw (583,22.4) node [anchor=north west][inner sep=0.75pt]    {$f_{1}^{3}$};
\draw (79,97.4) node [anchor=north west][inner sep=0.75pt]    {$1$};
\draw (135,97.4) node [anchor=north west][inner sep=0.75pt]    {$2$};
\draw (194,97.4) node [anchor=north west][inner sep=0.75pt]    {$3$};
\draw (335,250.4) node [anchor=north west][inner sep=0.75pt]    {$1.5$};
\draw (391,250.4) node [anchor=north west][inner sep=0.75pt]    {$3.5$};
\draw (605,92.4) node [anchor=north west][inner sep=0.75pt]    {$4$};

\end{tikzpicture}
\end{center}

For simplicity, we will assume all neurons have the same parameters $\alpha, \beta, \theta$. Let $w_1, w_2, w_3$ be the weights corresponding to the inputs to neurons 1, 2, and 3, respectively. Suppose that neuron 1 had firing times at $f_1^1 = 1, f^1_2 = 2,$ and $f^1_3 = 3$. Neuron 2 fired at $f^2_1 = 1.5$ and $f^2_2 = 3.5$. Finally neuron 3 fired at $f^3_1 = 4$. The input $x(t)$ causes neuron 1 to fire. Then note that the only firing times that could cause neuron 2 to fire at $f^2_1 = 1.5$ had to occur before $t = 1.5$. This is only $f^1_1 = 1$. After neuron 2 fires at $f^2_1$, its next firing time $f^2_2 = 3.5$ is affected by $f^1_1$, $f^1_2$ and $f^1_3$. And similarly, $f^2_1$ and $f^2_2$ affects $f^3_1$. This corresponds to the following causality diagram:
\begin{center}
\begin{tikzpicture}[x=0.5pt,y=0.5pt,yscale=-1,xscale=1]

\draw    (141.53,218.9) -- (141.53,146.19) ;
\draw [shift={(141.53,144.19)}, rotate = 90] [color={rgb, 255:red, 0; green, 0; blue, 0 }  ][line width=0.75]    (10.93,-3.29) .. controls (6.95,-1.4) and (3.31,-0.3) .. (0,0) .. controls (3.31,0.3) and (6.95,1.4) .. (10.93,3.29)   ;
\draw    (181.1,217.59) -- (390.26,140.95) ;
\draw [shift={(392.14,140.26)}, rotate = 159.88] [color={rgb, 255:red, 0; green, 0; blue, 0 }  ][line width=0.75]    (10.93,-3.29) .. controls (6.95,-1.4) and (3.31,-0.3) .. (0,0) .. controls (3.31,0.3) and (6.95,1.4) .. (10.93,3.29)   ;
\draw    (365.76,220.21) -- (437.59,152.12) ;
\draw [shift={(439.04,150.74)}, rotate = 136.53] [color={rgb, 255:red, 0; green, 0; blue, 0 }  ][line width=0.75]    (10.93,-3.29) .. controls (6.95,-1.4) and (3.31,-0.3) .. (0,0) .. controls (3.31,0.3) and (6.95,1.4) .. (10.93,3.29)   ;
\draw    (540.17,222.83) -- (500.08,148.57) ;
\draw [shift={(499.13,146.81)}, rotate = 61.64] [color={rgb, 255:red, 0; green, 0; blue, 0 }  ][line width=0.75]    (10.93,-3.29) .. controls (6.95,-1.4) and (3.31,-0.3) .. (0,0) .. controls (3.31,0.3) and (6.95,1.4) .. (10.93,3.29)   ;
\draw    (151.79,93.08) -- (255.44,54.46) ;
\draw [shift={(257.31,53.76)}, rotate = 159.56] [color={rgb, 255:red, 0; green, 0; blue, 0 }  ][line width=0.75]    (10.93,-3.29) .. controls (6.95,-1.4) and (3.31,-0.3) .. (0,0) .. controls (3.31,0.3) and (6.95,1.4) .. (10.93,3.29)   ;
\draw    (471.28,94.39) -- (379.31,53.26) ;
\draw [shift={(377.49,52.45)}, rotate = 24.09] [color={rgb, 255:red, 0; green, 0; blue, 0 }  ][line width=0.75]    (10.93,-3.29) .. controls (6.95,-1.4) and (3.31,-0.3) .. (0,0) .. controls (3.31,0.3) and (6.95,1.4) .. (10.93,3.29)   ;

\draw (112.63,236.97) node [anchor=north west][inner sep=0.75pt]    {$f_{1}^{1} =1$};
\draw (317.81,239.6) node [anchor=north west][inner sep=0.75pt]    {$f_{2}^{1} =2$};
\draw (514.2,235.66) node [anchor=north west][inner sep=0.75pt]    {$f_{3}^{1} =3$};
\draw (134.71,112.47) node [anchor=north west][inner sep=0.75pt]    {$f_{1}^{2} =1.5$};
\draw (433.69,109.84) node [anchor=north west][inner sep=0.75pt]    {$f_{2}^{2} =3.5$};
\draw (288.5,39.07) node [anchor=north west][inner sep=0.75pt]    {$f_{1}^{3} =4$};

\end{tikzpicture}
\end{center}
The arrows only point up to one level, which allows us to compute the necessary partial derivatives while computing the forward pass for the current layer (i.e., layer by layer). Note that while this simple example is for the reset to zero regime, where the membrane potential resets completely to 0 and all inputs in-between firing times accumulate until the next time the neuron fires, this kind of diagram can similarly be constructed for other regimes. For instance, if there is a time delay before inputs can start increasing the membrane potentials again, to decide the causal edges for a current firing time for a neuron we would have to look for input firing times that occurred at least ``time delay'' seconds after the current neuron's previous firing time.

We will use equations \eqref{eq:combined_response}, \eqref{eq:potential}, and \eqref{eq:potential_expanded} to define the following system. Since all neurons share the same parameters $\alpha, \beta$, we can simplify some notation and refer to the joint impulse response coming into a neuron as $h_{s+n}$ which corresponds to equation \eqref{eq:combined_response} and the impulse response for just the membrane potential dynamics as $h_n$ which corresponds to the $h_i^n$ term in equation \eqref{eq:potential}. Explicitly, 
\begin{align*}
    h_{s+n}(t) &= \frac{e^{-\alpha t} - e^{-\beta t} }{\beta - \alpha} u(t) \\
    h_n(t) &= e^{-\beta t} u(t)
\end{align*}
The firing time equations are explicitly given by the following:
\begin{align*}
w_1 \cdot \sum_{t: x(t) = 1 \wedge t < f^1_1} h_{s+n}(f^1_1 - t) &= \theta &\textnormal{Eq. for }f^1_1\\
w_1 \cdot \sum_{t: x(t) = 1 \wedge t < f^1_2} h_{s+n}(f^1_2 - t) - \theta \cdot h_n(f^1_2 - f^1_1) &= \theta &\textnormal{Eq. for }f^1_2\\
w_1 \cdot \sum_{t: x(t) = 1 \wedge t < f^1_3} h_{s+n}(f^1_3 - t) - \theta \cdot \big(h_n(f^1_3 - f^1_1) + h_n(f^1_3 - f^1_2)\big) &= \theta &\textnormal{Eq. for }f^1_3 \\
w_2 \cdot h_{s+n}(f^2_1 - f^1_1) &= \theta &\textnormal{Eq. for }f^2_1 \\
w_2 \cdot \big( h_{s+n}(f^2_2 - f^1_1) + h_{s+n}(f^2_2 - f^1_2) +  h_{s+n}(f^2_2 - f^1_3) \big) - \theta \cdot h_n(f^2_2 - f^2_1) &= \theta &\textnormal{Eq. for }f^2_2 \\
w_3 \cdot \big( h_{s+n}(f^3_1 - f^2_1) + h_{s+n}(f^3_1 - f^2_2)\big) &= \theta &\textnormal{Eq. for }f^3_1
\end{align*}

Now, all 6 equations are equations of the network weights ($w_1, w_2, w_3$) and the 6 firing times ($f^1_1, f^1_2, f^1_3, f^2_1, f^2_2, f^3_1$). Here, we invoke the implicit function theorem which will allow us to express firing times as a function of the weights.

We just need to check that the Jacobian of the above 6 equations (treated as a vector valued function) differentiated w.r.t. the 6 firing times is invertible. It turns out the causality structure will ensure that the Jacobian is always lower triangular once you sort by firing times. For feed-forward networks, this is also true if you sort by firing times by layer (since firing times within the same layer do not affect each other, and the firing times of deeper layers do not affect earlier ones). This Jacobian looks like

\begin{align*}
    \begin{blockarray}{ccccccc}
    & f^1_1 & f^1_2 & f^1_3 & f^2_1 & f^2_2 & f^3_1 \\
        \begin{block}{c(cccccc)}
        V_1(f^1_1)-\theta = 0 & x & & & & &   \\
        V_1(f^1_2)-\theta = 0 & x & x & & & &   \\
        V_1(f^1_3)-\theta = 0 & x & x & x & & &   \\
        V_2(f^2_1)-\theta = 0 & x & & & x & &   \\
        V_2(f^2_2)-\theta = 0 & x & x & x & x & x &   \\
        V_3(f^3_1)-\theta = 0 & & & & x & x & x  \\
        \end{block}
    \end{blockarray}
\end{align*}

where $x$ is marked for each equation there is a nontrivial derivative w.r.t. the corresponding variable. The lower triangular structure occurs because of the way later firing times cannot occur in the equations for earlier ones. 

Invertibility holds as long as the diagonal elements are non-zero. Since each equation is equal to the membrane potential at the firing threshold, the derivative of the membrane potential w.r.t. its firing time is the equal to the derivative of the membrane potential w.r.t. $t$ evaluated at the firing time, which is strictly positive because the potential is increasing at firing time. 

The Jacobian with respect to the network weights requires no special structure, but we can similarly calculate the partial derivatives of the above 6 equations now with respect to weights (to get a 6 by 3 matrix). Using \eqref{eq:ift_solve}, if we multiply the negative inverse of the Jacobian with respect to firing times (a 6 by 6 matrix) and the Jacobian with respect to weights (a 6 by 3 matrix), we get all partial derivatives of the 6 firing times with respect to all the weights (a 6 by 3 matrix). 

\subsection{Experiment details}\label{sec:experiment_details}
We include more details about our experiments here. 

\paragraph{XOR Task} The following hyperparameters of the 2-4-2 network were used for the XOR task. No special tuning of hyperparameters was used, and the network reliably converged to 100\% accuracy given the following parameters in Table \ref{tb:xor}.

\begin{table}[h]
\caption{Hyperparameters for XOR Task}
\label{tb:xor}
\begin{center}
\begin{tabular}{llr}
\textbf{SYMBOL}  &\textbf{DESCRIPTION} &\textbf{VALUE} \\
\hline \\
$\alpha = \frac{1}{\tau_s}$ & Inverse synaptic time constant & 1.0 \\
$\beta = \frac{1}{\tau_n}$ & Inverse membrane time constant & 0.99 \\
$\theta$ & Threshold & 1.0 \\
$T$ & Maximum time & 2.0 \\
$t_{\textnormal{early}}$ & Minimum time & 0.0 \\
  & Hidden sizes & [4] \\
  & Hidden weights mean & [3.0] \\
  & Hidden weights stdev & [1.0] \\
  & Output weights mean & 2.0 \\
  & Ouput weights stdev & 0.1 \\
  & Optimizer & Adam \\
$\beta_1$ & Adam parameter & 0.9 \\
$\beta_2$ & Adam parameter & 0.999 \\
$\epsilon$ & Adam parameter & $1e-8$ \\
$\eta$ & Learning rate & 0.1 \\
$\gamma$ & Regularization factor & 0.2 \\
$\tau_0$ & First loss time constant & 0.1 \\
$\tau_1$ & Second loss time constant & 1.0 \\
\end{tabular}
\end{center}
\end{table}

\paragraph{Iris Dataset} We did a systemic grid search over hyperparameters to find a network suitable for the Iris classification. There were several networks which achieved 100\% test accuracy. One such set of hyperparameters is given in Table \ref{tb:iris}.

\begin{table}[h]
\caption{Hyperparameters for Iris Classification}
\label{tb:iris}
\begin{center}
\begin{tabular}{llr}
\textbf{SYMBOL}  &\textbf{DESCRIPTION} &\textbf{VALUE} \\
\hline \\
$\alpha = \frac{1}{\tau_s}$ & Inverse synaptic time constant & 1.0 \\
$\beta = \frac{1}{\tau_n}$ & Inverse membrane time constant & 0.9 \\
$\theta$ & Threshold & 1.0 \\
$T$ & Maximum time & 16.0 \\
$t_{\textnormal{early}}$ & Minimum time & 0.0 \\
  & Hidden sizes & [10] \\
  & Hidden weights mean & [3.0] \\
  & Hidden weights stdev & [1.0] \\
  & Output weights mean & 2.0 \\
  & Ouput weights stdev & 0.1 \\
  & Optimizer & Adam \\
$\beta_1$ & Adam parameter & 0.9 \\
$\beta_2$ & Adam parameter & 0.999 \\
$\epsilon$ & Adam parameter & $1e-8$ \\
$\eta$ & Learning rate & 0.05 \\
$\gamma$ & Regularization factor & 0.1 \\
$\tau_0$ & First loss time constant & 1.0 \\
$\tau_1$ & Second loss time constant & 1.0 \\
\end{tabular}
\end{center}
\end{table}

\paragraph{Yin-Yang Dataset} The following Table \ref{tb:hyperparams} describes the hyperparameters used for training the SNN on the Yin-Yang dataset. The hyperparameters were chosen by a manual search through several combinations of the architecture and the parameters shown in the table. The final experiments were done on machines part of an internal cluster with 48 CPU and 5 GB memory, which results in training over 300 full epochs through the entire training dataset of 5000 examples and evaluation on the entire test dataset of 1000 examples completing in approximately 3 hours. 

\begin{table}[h]
\caption{Hyperparameters for Yin-Yang Simulations} \label{tb:hyperparams}
\begin{center}
\begin{tabular}{llr}
\textbf{SYMBOL}  &\textbf{DESCRIPTION} &\textbf{VALUE} \\
\hline \\
$\alpha = \frac{1}{\tau_s}$ & Inverse synaptic time constant & 0.999 \\
$\beta = \frac{1}{\tau_n}$ & Inverse membrane time constant & 1.0 \\
$\theta$ & Threshold & 1.0 \\
$T$ & Maximum time & 2.0 \\
$t_{\textnormal{early}}$ & Minimum time & 0.15 \\
$t_{\textnormal{bias}}$ & Bias time & 0.9 \\
  & Hidden sizes & [150] \\
  & Hidden weights mean & [1.5] \\
  & Hidden weights stdev & [0.8] \\
  & Output weights mean & 2.0 \\
  & Ouput weights stdev & 0.1 \\
  & Minibatch size & 150 \\
  & Epochs & 300 \\
  & Optimizer & Adam \\
$\beta_1$ & Adam parameter & 0.9 \\
$\beta_2$ & Adam parameter & 0.999 \\
$\epsilon$ & Adam parameter & $1e-8$ \\
$\eta$ & Learning rate & 0.0005 \\
$\gamma$ & Regularization factor & 0.005 \\
$\tau_0$ & First loss time constant & 0.2 \\
$\tau_1$ & Second loss time constant & 1.0 \\
\end{tabular}
\end{center}
\end{table}

\end{document}